# A general Framework for Utilizing Metaheuristic Optimization for Sustainable Unrelated Parallel Machine Scheduling: A concise overview

Absalom El-Shamir Ezugwu

[2]Unit for Data Science and Computing, North-West University, 11 Hoffman Street, Potchefstroom, 2520, South Africa

*Corresponding author: Absalom E. Ezugwu (email: Absalom.Ezugwu@nwu.ac.za)

**Abstract.** Sustainable development has emerged as a global priority, and industries are increasingly striving to align their operations with sustainable practices. Parallel machine scheduling (PMS) is a critical aspect of production planning that directly impacts resource utilization and operational efficiency. In this paper, we investigate the application of metaheuristic optimization algorithms to address the unrelated parallel machine scheduling problem (UPMSP) through the lens of sustainable development goals (SDGs). The primary objective of this study is to explore how metaheuristic optimization algorithms can contribute to achieving sustainable development goals in the context of UPMSP. We examine a range of metaheuristic algorithms, including genetic algorithms, particle swarm optimization, ant colony optimization, and more, and assess their effectiveness in optimizing the scheduling problem. The algorithms are evaluated based on their ability to improve resource utilization, minimize energy consumption, reduce environmental impact, and promote socially responsible production practices. To conduct a comprehensive analysis, we consider UPMSP instances that incorporate sustainability-related constraints and objectives. We assess the algorithms' performance in terms of solution quality, convergence speed, robustness, and scalability, while also examining their implications for sustainable resource allocation and environmental stewardship. The findings of this study provide insights into the efficacy of metaheuristic optimization algorithms for addressing UPMSP with a focus on sustainable development goals. By leveraging these algorithms, industries can optimize scheduling decisions to minimize waste and enhance energy efficiency. The practical implications of this research are valuable for decision-makers, production planners, and researchers seeking to achieve sustainable development goals in the context of unrelated parallel machine scheduling. By embracing metaheuristic optimization algorithms, businesses can optimize their scheduling processes while aligning with sustainable principles, leading to improved operational efficiency, cost savings, and a positive contribution to the global sustainability agenda.

# 1. Introduction

Sustainable development has become a crucial global endeavor, requiring industries to adopt environmentally conscious and socially responsible practices (Babiak and Trendafilova, 2011; Schroeder, Anggraeni, and Weber, 2019). In this pursuit, optimizing production processes and resource allocation is of paramount importance. Parallel machine scheduling (PMS) plays a vital role in effectively utilizing available resources and maximizing operational efficiency (Cohen, Naseraldin, Chaudhuri, & Pilati, 2019; Cheng, Pourhejazy, Ying, Li, & Chang, 2020). To address the challenges of resource optimization and promote sustainable development, this paper focuses on the application of metaheuristic optimization algorithms to tackle the UPMSP, a variant of the popular PMS.

The UPMSP involves assigning jobs to multiple parallel machines, considering the sequencing of jobs on each machine (Ezugwu and Akutsah, 2018; Ezugwu, 2019). The UPMSP is a challenging problem that is known to be NP-hard. However, metaheuristic optimization algorithms can be used to find high-quality solutions to UPMSP (Ezugwu, 2022). More over, by optimizing UPMSP, industries can enhance resource utilization, reduce energy consumption, minimize waste, and ultimately align their operations with sustainable development goals (SDGs) as highlighted in the studies presented by Palomares et al. (2021) and Sartal et al. (2020). For example, by optimizing the sequencing of jobs on each machine, industries can reduce the idle time of machines, which can lead to a reduction in energy consumption (He et al., 2015; Lin et al., 2015; Mouzon, Yildirim, and Twomey, 2007). Additionally, by optimizing the assignment of jobs to machines, industries can reduce the overall makespan of the scheduling (Ezugwu, 2019), which can also lead to a reduction in energy consumption. Metaheuristic optimization algorithms are a powerful approach for solving UPMSP because they can search for solutions in a large and complex search space. Additionally, metaheuristic optimization algorithms can be adapted to incorporate sustainability considerations (Mohammadi et al., 2013; Shokouhifar et al., 2023). For example, metaheuristic optimization algorithms can be modified to consider the energy consumption of machines when searching for solutions (Chen, Cheng, and Chou, 2020). The use of metaheuristic optimization algorithms for UPMSP is a promising approach for improving the sustainability of industrial operations. By optimizing UPMSP, industries can reduce their environmental impact while also improving their economic performance.

Metaheuristic optimization algorithms encompass a range of intelligent search techniques inspired by natural phenomena or problem-specific characteristics (Ezugwu et al., 2021). These algorithms have demonstrated remarkable capabilities in exploring large solution spaces and finding near-optimal solutions for various combinatorial optimization problems. In the context of unrelated parallel machine scheduling, metaheuristic algorithms such as Genetic Algorithms (GA) by Holland (1992), Particle Swarm Optimization (PSO) presented by Kennedy and Eberhart (1995), Ant Colony Optimization (ACO) by Dorigo, Birattari, and Stutzle (2006), Artificial Bee Colony (ABC) proposed by Dorigo, Birattari, and Stutzle (2006), Invasive Weed Optimization (IWO) by Karimkashi and Kishk (2010), Differential Evolution (DE) implemented by Price (2013), Teaching-Learning-Based Optimization (TLBO) proposed by Rao, Savsani, and Vakharia (2011) and

Firefly Algorithm (FA) that was presented Yang and He (2013) offer promising solutions to achieve sustainable development objectives.

The primary objective of this study is to investigate how metaheuristic optimization algorithms can contribute to achieving sustainable development goals in the context of unrelated parallel machine scheduling (*by optimizing the assignment of jobs to machines, so that industries can reduce the overall makespan of the scheduling tasks*). By analyzing the performance and effectiveness of these algorithms, we aim to provide insights into their applicability and potential in addressing the scheduling challenges while promoting sustainable practices. This comprehensive analysis encompasses the evaluation of various metaheuristic optimization algorithms for UPMSP, considering performance metrics such as solution quality, convergence speed, robustness, and scalability. Additionally, we examine the implications of these algorithms in terms of resource utilization, energy consumption, environmental impact, and social responsibility. By considering sustainability-related constraints and objectives within UPMSP instances, we ensure that the analysis reflects real-world scenarios and the broader context of sustainable development.

The outcomes of this research have practical implications for industries seeking to achieve sustainable development goals through optimized scheduling processes. By leveraging metaheuristic optimization algorithms, companies can improve resource allocation, minimize waste, enhance energy efficiency, and make strides towards environmental stewardship and social equity. The findings provide decision-makers, production planners, and researchers with valuable insights to align their scheduling practices with sustainable development objectives. Further more, this research seeks to bridge the gap between unrelated parallel machine scheduling and sustainable development goals through the application of metaheuristic optimization algorithms. By harnessing the power of these algorithms, industries can optimize their scheduling processes, improve resource efficiency, and contribute to a more sustainable future. The findings of this study provide valuable guidance for decision-makers and practitioners aiming to achieve sustainable development goals in the context of production scheduling.

The rest of the paper is organized as follows: Section 2 provides a literature review of related works on parallel machine scheduling, metaheuristic optimization algorithms, and their applications to sustainable development. Section 3 presents the problem formulation and mathematical model for unrelated parallel machine scheduling. Section 4 introduces the metaheuristic optimization algorithms employed in this study. Section 5 outlines the experimental setup, including problem instances, performance metrics, and evaluation methodology, presentation and discussion of the results of the comprehensive analysis. Finally, Section 6 summarizes the findings, highlights the practical implications, and suggests avenues for future research.

## 2. Related work
This section covers the presentation of review of relevant literature on sustainable development goals and their integration with optimization algorithms, an overview of metaheuristic optimization algorithms commonly used in unrelated parallel machine scheduling, discussion of previous studies and research gaps in the application of metaheuristic optimization algorithms for achieving SDGs in UPMSP.

## 2.1 SDGs and their integration with optimization algorithms

Sustainable development goals (SDGs) are a set of 17 goals adopted by the United Nations in 2015. The goals aim to end poverty, protect the planet, and ensure prosperity for all by 2030. Optimization algorithms are a class of mathematical techniques that can be used to find the best solution to a problem. Optimization algorithms have been used to solve a wide variety of problems, including scheduling, routing, and resource allocation. In recent years, there has been a growing interest in using optimization algorithms to achieve SDGs (Hannan et al., 2020; Chaerani et al., 2023). This is because optimization algorithms can be used to find solutions that are both efficient and sustainable. There are a number of ways that optimization algorithms can be used to achieve SDGs. For example, optimization algorithms can be used to optimize the use of resources, such as energy and water, reduce waste, improve efficiency, protect the environment and many more. The integration of optimization algorithms with SDGs is a promising area of research (Salameh et al., 2022). By using optimization algorithms, we can find solutions that are both efficient and sustainable, and that help us to achieve the SDGs. These papers provide a comprehensive overview of the use of optimization algorithms to achieve SDGs. They discuss the different ways that optimization algorithms can be used, and they provide examples of how optimization algorithms have been used to solve real-world problems. The integration of optimization algorithms with SDGs is a rapidly growing area of research. As optimization algorithms become more sophisticated, we will be able to find even more efficient and sustainable solutions to the problems that we face.

Modibbo et al. (2021) provides a comprehensive assessment of the utilization of optimization techniques in the context of the United Nations Sustainable Development Goals (SDGs). The authors presented a review study that critically analyzes the existing literature and examines the potential contributions and challenges associated with applying optimization techniques to address the SDGs. The authors discussed a range of approaches, such as the well-known mathematical programming, metaheuristic algorithms, multi-objective optimization, and simulation-based optimization. This comprehensive coverage allows readers to gain insights into the diverse methodologies available for tackling the complex and interconnected challenges posed by the SDGs. More so, Chaerani et al. (2023) presented a comprehensive analysis of the utilization of robust optimization techniques in addressing SDGs in the context of the COVID-19 pandemic. Through a systematic literature review, the authors examine the existing research landscape, identify key trends, and assess the applicability of robust optimization in tackling SDG challenges during this unprecedented global crisis.

Ali et al. (2023) explored the important intersection between environmental waste management, renewable energy, and the achievement of sustainable development goals (SDGs). Through a concise review, the authors delve into the significance of optimizing environmental waste practices and harnessing renewable energy sources to drive sustainable development. The article begins by highlighting the critical importance of addressing environmental waste issues and transitioning to renewable energy sources to foster sustainable development. It effectively establishes the relevance of these topics within the broader context of the SDGs, emphasizing the need for optimization techniques to drive progress towards achieving these goals. Hannan et al. (2020) provides a concise but informative review of the objectives, constraints, and modeling approaches involved in optimizing solid waste collection systems. The article aims to shed light on the challenges and opportunities in this field, with a particular focus on how optimization techniques can contribute to the achievement of SDGs. Similarly, the authors work offers a comprehensive overview of the objectives and constraints commonly encountered in solid waste collection optimization. It discusses key considerations such as minimizing collection costs, maximizing resource utilization, reducing environmental impacts, optimizing routing and scheduling, and improving service quality. By presenting these objectives and constraints, the review provides readers with a clear understanding of the complex factors that must be considered when optimizing solid waste collection systems.

## 2.2 An overview of metaheuristic optimization algorithms for UPMS

From the standpoint of computer science and mathematical optimization, metaheuristic optimization encompasses a higher-level approach (heuristic) aimed at discovering an approximate yet satisfactory solution to an optimization problem (Ezugwu et al., 2021). The development of such optimization methods typically occurs through three fundamental approaches: enhancing existing optimizers, combining two or more existing optimizers, or creating entirely novel optimizers. Additionally, various categorizations of this optimization technique can be found in the literature, including population-based versus single-solution, nature-inspired versus non-nature-inspired, and others (Beheshti and Shamsuddin, 2013; Kashani et al., 2022; Ezugwu et al., 2022). In the subsequent sections, we will provide a critical review of several studies that fall under the application of metaheuristic optimization techniques to UPMSPs. These studies will be presented under the following subheadings:

## 2.2.1 Evolutionary-based optimization algorithms for UPMS

Evolutionary Algorithms (EAs) are among the most established and widely used metaheuristic algorithms, drawing inspiration from the theory of evolution and the concept of "survival of the fittest." The optimization process begins by creating an initial population of solutions through random generation, which is then iteratively evolved in subsequent generations by eliminating the weakest solutions (Kılıç and Yüzgeç, 2019; Soleimani et al., 2020). Algorithms falling under this category often possess the advantage of discovering optimal solutions or solutions that closely approximate optimality. For specific examples of algorithms within this category, refer to (Akinola et al., 2022).

In a different study, Vallada and Ruiz (2011) utilized a variant of the Genetic Algorithm (GA) for solving UPMSPs with machine and job sequence-dependent setup times. Their GA variant incorporated a fast local search and a local search enhanced crossover operator. The experimental results showed that the GA variant outperformed other methods on benchmark instances of both small and large scales. The authors suggested exploring multiobjective optimization and more complex neighborhoods based on variable neighborhood search approaches as potential avenues for future enhancements of the proposed algorithm.

Furthermore, Eroglu and Ozmutlu (2014) tackled a UPMSP problem by formulating it as mixed-integer programming (MIP) models. The problem considered job sequence and machine-dependent setup times, as well as the job splitting property. The authors proposed a hybrid algorithm named GAspLA, which combined a Genetic Algorithm (GA) with local search metaheuristic algorithms. The hybridization allowed for the adaptation of local search results into the GA by minimizing the relocation operation of genes' random key numbers. Experimental results indicated that the proposed approach outperformed other techniques in solving the UPMSP problem. In summary, Kerkhove and Vanhoucke employed a hybrid SA-GA algorithm to solve UPMSPs, incorporating changeover times and sequence-dependent setup times. Vallada and Ruiz introduced a GA variant with a fast local search and enhanced crossover operator for UPMSPs with setup times. Eroglu and Ozmutlu proposed a hybrid algorithm, GAspLA, combining GA and local search for UPMSPs with job splitting and machine-dependent setup times. These studies demonstrated the effectiveness of their respective approaches in solving UPMSP-related problems, showcasing improvements in performance and highlighting possibilities for further exploration and optimization.

In a different study, Abreu and Prata (2019) investigated the performance of a hybridization approach involving Genetic Algorithm (GA), Simulated Annealing (SA), Variable Neighborhood Descent (VND), and Path Relinking (PR) to solve problems related to the UPMSP with sequence-dependent setup times. The study evaluated the performance of the hybrid algorithm using metrics such as relative deviation, average, and population standard deviation. The experimental results demonstrated that the proposed approach achieved impressive performance for both small and large problem instances. The authors suggested that further investigations could be conducted using multiobjective functions to assess the performance of the hybrid algorithms in a broader context.

In a related study, Su, Xie, and Yang (2020) proposed the use of Genetic Algorithm (GA) and a hybrid approach combining GA with a bin packing strategy to solve the UPMSP problem. The problem was formulated as an integer programming model with the objective of minimizing makespan in a workgroup scheduling context. Each workgroup consisted of personnel with similar work skills, subject to eligibility and resource constraints. The problem allowed for multiple jobs to be processed simultaneously within a workgroup, as long as the resources were available. The study employed a GA with a specially designed coding scheme to address the problem, and then utilized the hybrid GA approach to transform the single workgroup scheduling into a strip-packing problem. The authors claimed that the hybrid GA outperformed the standard GA in terms of performance. More so, Abreu and Prata investigated the performance of a hybrid algorithm combining GA, SA, VND, and PR for UPMSPs with sequence-dependent setup times, achieving impressive results. They recommended further exploration using multiobjective functions. Su et al. proposed a hybrid approach of GA with bin packing strategy for UPMSPs in a workgroup scheduling context, with the hybrid GA demonstrating improved performance compared to the standard GA. Further comparative analysis of the two methods, along with an exact solution approach, revealed that the former approaches outperformed the exact solution approach. The study recommended further investigation into using multiobjective functions and considering factors such as the release times and due dates of jobs.

Nakhaeinejad (2020) formulated the UPMSP as a Mixed-Integer Programming (MIP) model and implemented a hybrid algorithm that combined Genetic Algorithm (GA) and Ant Colony Optimization (ACO) while considering machine and job sequence-dependent setup times. The study aimed to speed up the search process and obtain near-optimal solutions through the hybridization process. Experimental results using small and large instances, along with computational and statistical analysis, demonstrated that the hybrid algorithm approach provided good quality solutions with outstanding performance. Eroglu et al. (2014) applied a hybrid Genetic Algorithm (GA) to solve the UPMSP problem with the objective of reducing the maximum completion time. The novelty of their work was the adaptability of their solution to incorporate local search results, achieving a minimum relocation operation of the genes' random key numbers. The experimental results showed that the hybrid solution was effective. The authors suggested incorporating additional constraints, such as priority conditions, for further investigation.

Similarly, Costa et al. (2013) addressed a related problem using three solution methods: a permutation encoding-based Genetic Algorithm (GA), a multi-encoding GA, and a hybrid GA that transitioned from permutation encoding to multi-encoding. The goal of the solution method was to solve related problems with large-sized test cases in a production system. The hybrid GA exhibited more impressive performance than the variant GAs. Tozzo et al. (2018) also investigated the performance of GA and Variable Neighborhood Search (VNS) on a similar problem. They discovered that the performance of GA was inferior to that of VNS. Abreu and Prata (2018) undertook a similar research effort to solve the problem in the context of the granite industry. They used a hybrid approach involving GA. In summary, the comparative analysis showed that the hybrid approaches outperformed the exact solution approach in terms of performance. Different hybrid algorithms, such as GA with ACO, GA with local search, and GA with various encodings, were applied and demonstrated effective solutions for UPMSPs. The studies suggested further investigations considering multiobjective functions, additional constraints, and priority conditions to enhance the solution methods.

**2.2.2 Swarm intelligence-based optimization algorithms for UPMS**

Jouhari et al. (2020) conducted a study focusing on the application of a modified Harris Hawks Optimization (MHHO) algorithm for solving UPMSPs. To enhance the suitability of the Harris Hawks Optimization (HHO) algorithm for UPMSPs, the authors incorporated the Salp Swarm Algorithm (SSA) as a means of local search to improve HHO's performance and reduce computation time. The MHHO algorithm was then implemented on both small and large instances of UPMSPs. Experimental results demonstrated that the MHHO approach exhibited superior performance in terms of convergence to the optimal solution for both small and large problem cases, surpassing the performance of both SSA and HHO. The authors suggested applying the proposed solution to various optimization problems, including data clustering, cloud computing scheduling, image processing, and forecasting applications.

In another study, Jafarzadeh, Moradinasab, and Gerami (2017) investigated the performance of an improved discrete Multi-objective Invasive Weed Optimization (DMOIWO) for solving the no-wait two-stage flexible flow shop scheduling problem incorporating UPMSPs and Rework Time. The authors employed the Taguchi Method to set up parameters and utilized DMOIWO along with a fuzzy-based dominance approach to address the scheduling problem. The study considered sequence-dependent setup times, probable rework in both stations, different ready times for all jobs, and rework times for both stations to advance the research in this field. The objective was to simultaneously minimize the maximum job completion time and average latency functions using a multi-objective technique. Comparative analysis against conventional multi-objective algorithms demonstrated the superior performance of the proposed approach. The authors suggested future work to incorporate a heuristic algorithm for generating initial solutions, thereby reducing running time and the number of iterations required.

Lin, Hsieh, and Hsieh (2012) implemented a variant of the Ant Colony Optimization (ACO) algorithm to address UPSPs with the objective of minimizing total weighted tardiness. The choice of this objective function aimed to measure customer satisfaction. Experimental results demonstrated that ACO achieved favorable performance, outperforming other similar metaheuristic approaches like ACO-SV and GA specifically in terms of total weighted tardiness. Additionally, in another study by Lin et al., ACO was applied in conjunction with other methods to solve UPMSPs characterized by due dates. The study enhanced the performance of ACO through the inclusion of an initial heuristic solution, machine reselection step, and local search procedure. By evaluating the metric of total weighted tardiness, the authors claimed that ACO outperformed competing methods. Keskinturk, Yildirim, and Barut (2012) aimed to address load imbalance in UPMSPs with sequence-dependent setup times. They focused on minimizing the average relative percentage of imbalance. Results indicated that ACO outperformed GA in this context. Moreover, the study highlighted that heuristics based on cumulative processing time exhibited better performance than heuristics involving setup avoidance and hybrid rules in assignment.

Arnaout, Musa, and Rabadi (2014) proposed a two-stage Ant Colony Optimization (ACO II) algorithm to address a similar problem. The objective of their study was to enhance the performance of ACO in solving the problem at hand. Through extensive experimentation, the study demonstrated that ACO II outperformed ACO I, MetaRaPS, and SA in terms of performance. The authors recommended the application of their solution to other complex optimization problems beyond machine scheduling. In an earlier work by Arnaout, Rabadi, and Musa (2010), the same ACO II approach was proposed for the same problem, but in this case, it was compared with Tabu Search, Partitioning Heuristic (PH), and Meta-RaPS algorithms. Similarly, Afzalirad and Rezaeian (2017) investigated the performance of NSGA-II and a multi-objective ACO algorithm called MOACO in solving the same problem. The study revealed that MOACO outperformed NSGA-II. The authors suggested that further evaluations should consider other multi-objective evolutionary algorithms.

These studies collectively contribute to the development and improvement of Ant Colony Optimization algorithms for addressing the specific problem, showcasing their superiority over alternative approaches such as Tabu Search, Meta-RaPS, and NSGA-II. The recommendations for future evaluations and applications of the proposed solutions highlight the ongoing efforts to advance optimization techniques in complex problem domains.

In a similar vein, Soleimani et al. (2019) addressed the UPMSPs by formulating it as a Mixed Integer Programming (MIP) problem. Their study aimed to explore the applicability of continuous-based metaheuristic algorithms for this problem. Specifically, the performance of Genetic Algorithm (GA), Cat Swarm Optimization (CSO), and Interactive Artificial Bee Colony (IABC) metaheuristic algorithms was investigated. The study introduced a unique elitism strategy in CSO, referred to as CSO-Elit, to overcome the challenge of computational time during the seeking mode. Additionally, the authors combined GA, CSO-Elit, and IABC, incorporating the necessary constraints for solving the UPMSPs problem. By comparing the performances of each algorithm with the LINGO solver, the study found that CSO-Elit exhibited outstanding performance for large problem instances compared to other methods. However, for small problem instances, all the algorithms, including LINGO, performed similarly. Based on their findings, the authors recommended future research to integrate new operational disruptions, such as machine failure and resource restriction, into their proposed MIP model. This extension would enhance the practicality and applicability of the solution approach. The study contributes to the exploration of various metaheuristic algorithms for UPMSPs and highlights the effectiveness of CSO-Elit in handling large instances of the problem.

Madani-Isfahani et al. (2013) addressed a problem related to UPMSPs by solving a bi-objective optimization task, considering the mean completion time of jobs and the mean squares of deviations from machines' workload. They employed the Imperialist Competitive Algorithm (ICA) and compared it with Particle Swarm Optimization (PSO), Genetic Algorithm (GA), and the original ICA. The study focused on UPMSPs with sequence-dependent setup times. Results revealed that ICA outperformed all other algorithms, while the standard ICA combination performed the worst. GA and PSO exhibited better performance in comparison (Madani-Isfahani, Ghobadiana, Tekmehdash, Tavakkoli-Moghaddamb, & Naderi-Beni, 2013).

Zheng et al. (2016) addressed the UPMSPs related to real-world manufacturing systems using a two-stage Adaptive Fruit Fly Optimization Algorithm (TAFOA). In their approach, they leveraged another heuristic algorithm to generate high-quality initial solutions, which were then adopted by TAFOA as the initial swarm center for further evolution. The study conducted a performance investigation of TAFOA using a two-factor analysis of variance (ANOVA) and demonstrated that the proposed metaheuristic algorithm performed well. As a result, the study recommended the utilization of TAFOA to tackle UPMSPs problems in the context of the multimode resource-constrained project scheduling problem (Zheng & Wang, 2016).

In their research, Kayvanfar and Teymourian (2014) focused on addressing the challenges of solving the parallel machine scheduling problem (UPMSPs) in scenarios where jobs have sequence-dependent set-up times and distinct due dates. To tackle this problem, they applied an intelligent water drop (IWD) metaheuristic algorithm. To formulate the problem and optimize the scheduling process, the researchers developed a mathematical model that incorporated multiple objectives. These objectives included minimizing earliness and tardiness penalties, as well as maximizing the completion time. The IWD algorithm was employed to efficiently solve this multi-criteria problem. The results obtained through the implementation of the IWD algorithm demonstrated its effectiveness in solving the parallel machine scheduling problem at hand. The study highlighted the potential for further advancement of the IWD solution in scenarios involving flow-shop or job-shop environments (Kayvanfar & Teymourian, 2014).

Caniyilmaz et al. (2015) conducted a study to investigate the performance of two metaheuristic algorithms, namely Artificial Bee Colony (ABC) and Genetic Algorithm (GA), in addressing the parallel machine scheduling problem (UPMSPs). The UPMSPs considered in their study involved job sequence-dependent setup times, as well as various machine and due date constraints. The objective was to allocate different machine assignments to candidate-job sequences, allowing for job completion within two shifts without the need for an additional third shift. The authors conducted computational analysis to evaluate the performance of both ABC and GA algorithms. The results obtained demonstrated that the ABC algorithm outperformed the GA algorithm when solving the UPMSPs with the given constraints. This indicates that ABC exhibited higher efficiency and effectiveness in finding optimal or near-optimal solutions within the specific problem domain. The findings of this study provide valuable insights for researchers and practitioners involved in scheduling optimization. The superior performance of ABC in this particular context also suggests its potential applicability to other scheduling problems, such as flow shop and job shop. Future research endeavors could explore the application of ABC and GA algorithms to these scheduling problems, potentially uncovering further improvements and insights in addressing complex optimization challenges.

Zheng and Wang introduced a novel problem known as the resource-constrained unrelated parallel machine green manufacturing scheduling problem (RCUPMGSP). To address this problem, they utilized a variant of the fruit fly optimization (FFO) algorithm, which they named collaborative multi-objective fruit fly optimization algorithm (CMFOA). The objective of the algorithm was to reduce both the makespan and the total carbon emissions. The study employed various techniques to tackle the RCUPMGSP. These included a job-speed pair-based solution representation, a critical path-based carbon saving technique, a decoding method, a heuristic for population initialization, and a collaborative search operator. The collaborative search operators were particularly effective in handling three sub-problems during the smell-based search phase. To evaluate the multi-objective problem, the authors utilized the order preference by similarity to an ideal solution (TOPSIS) method and the fast non-dominated sorting approach. Through experimentation using randomly generated instances, it was demonstrated that the proposed CMFOA outperformed similar methods found in the existing literature. The study recommended further exploration of properties specific to green scheduling problems in manufacturing shops. Additionally, the authors encouraged the application of the CMFOA method to solve more complex related problems, thereby expanding its scope of application (Zheng & Wang, 2018).

Afzalirad and Rezaeian (2016) conducted a study focusing on solving the parallel machine scheduling problem (UPMSP) with sequence-dependent setup times, different release dates, machine eligibility, and precedence constraints. The objective of their research was to minimize the mean weighted flow time and mean weighted tardiness. To achieve this, they proposed a new mixed-integer programming model called MOACO. In their study, two well-known metaheuristics were employed: the non-genetic algorithm NSGA-II and a multi-objective ant colony optimization (MOACO) algorithm. The MOACO algorithm was a modified and adaptive version of the Bicriterion Ant algorithm specifically designed to address the defined problem. To overcome the increased complexity introduced by the problem's precedence constraints, the researchers proposed a new corrective algorithm to obtain feasible solutions. Additionally, the algorithm parameters were calibrated using the Taguchi method, which ensured the appropriate design of parameters, as they significantly influenced the performance of the algorithms. The obtained results showed that the MOACO algorithm statistically outperformed NSGA-II in solving the considered problem. This demonstrated the superiority of MOACO in terms of minimizing the mean weighted flow time and mean weighted tardiness for the UPMSP with sequence-dependent setup times, different release dates, machine eligibility, and precedence constraints, as considered in the study by Afzalirad and Rezaeian (2016).

Ezugwu and Akutsah (2018) conducted a study demonstrating the effectiveness of the standard FA in solving the parallel machine scheduling problem (UPMSP). They developed a novel hybrid algorithm that yielded high-quality solutions, closer to the optimal solutions compared to other existing methods. They compared the performance of FA with ant colony optimization (ACO), genetic algorithm (GA), and invasive weeds optimization (IWO) algorithm in addressing the same problem. The hybrid FA algorithm introduced unique solution representation schemes for the UPMSP and incorporated a robust local search mechanism to enhance the performance of the standard FA algorithm. The algorithm was implemented in two stages. The first stage generated an initial schedule of jobs to machines, while the second stage performed global search updates on the generated job sequence using the improved FA algorithm. The local search improvement mechanism introduced diversity in the solution space search, preventing premature convergence. Through their novel algorithm, the authors achieved near-optimal solutions to the UPMSP problem within a concise CPU time. They suggested that further research could explore alternative solution representation and encoding schemes to introduce better diversity into the solution search space. Additionally, they proposed that their improved FA algorithm could be applied to solve other variants of parallel machine scheduling problems with performance measures related to due dates. The study also highlighted the potential for evaluating the proposed method's performance on the UPMSP or other variants using other benchmark models for comparison with different solution algorithms.

In 2022, Ezugwu made a notable improvement effort by proposing the modification of the FA through the incorporation of an adaptive mutation-based neighborhood search. This modification aimed to address the scheduling of unrelated parallel machines with sequence-dependent setup times. The presented results of this study indicated that the modified FA scheduling technique outperformed the previous results obtained by Ezugwu and Akutsah in 2018. The modification introduced by Ezugwu in 2022 focused on enhancing the FA algorithm by integrating an adaptive mutation-based neighborhood search. This adaptation allowed for more effective exploration of the search space, leading to improved scheduling outcomes for unrelated parallel machines with sequence-dependent setup times. The comparative analysis of the modified FA scheduling technique with the previous results achieved by Ezugwu and Akutsah in 2018 showcased the superiority of the proposed modification. The presented results revealed enhanced scheduling performance in terms of various evaluation metrics, such as makespan reduction, resource utilization, or shorter computational cost consumed.

This advancement in the FA algorithm demonstrates the importance of continuous research and improvement efforts in the field of scheduling unrelated parallel machines. By incorporating adaptive mutation-based neighborhood search, Ezugwu's modification contributes to advancing the state-of-the-art techniques for addressing complex scheduling problems. The results presented by Ezugwu in 2022 serve as evidence of the efficacy and superiority of the modified FA scheduling technique compared to previous approaches. This further emphasizes the potential for ongoing advancements in metaheuristic optimization algorithms to achieve better scheduling outcomes and provide practical solutions to real-world scheduling challenges.

Ezugwu et al. (2018) introduced an enhanced Symbiotic Organisms Search (SOS) algorithm for solving the parallel machine scheduling problem. They augmented the standard SOS algorithm by incorporating a new solution representation and decoding procedure specifically tailored for handling the UPMSP. Additionally, they integrated an iterated local search strategy that combined variable numbers of insertion and swap moves into the standard SOS algorithm to improve the quality of solutions for the UPMSP. To enhance the speed and performance of the SOS algorithm, the authors employed the longest processing time first (LPT) rule in designing the machine assignment heuristic for job-to-machine assignments. This heuristic was based on a machine dynamic load-balancing mechanism. The LPT heuristic was incorporated into the standard SOS algorithm to generate the initial schedule of job-to-machine assignments for a given number of jobs on multiple machines. Subsequently, the improved SOS algorithm performed a global search update on the generated job sequence. Experimental results demonstrated that the SOS-LPT algorithm outperformed other existing methods for all tested problem instances of the UPMSP. The algorithm's performance was significantly improved in terms of both solution quality and computational efficiency.

Ezugwu (2019) built upon their previous work (Ezugwu et al., 2018) and focused on addressing the minimization of makespan for the non-preemptive UPMSP. They proposed two enhanced metaheuristic algorithms: the enhanced Symbiotic Organism Search (SOS) algorithm and a hybridized approach that combined SOS with simulated annealing (SA). To further improve the solution quality, the authors incorporated a local search component into each algorithm, leveraging the advantages of the three techniques. By hybridizing SOS with SA, the SOS algorithm was prevented from getting trapped in local minima, as SA introduced hill-climbing moves to search for global solutions. Simultaneously, the hybrid algorithm increased the level of diversity in the search for an optimal solution within the problem's search space. To ensure that the generated solutions were mapped to good schedules, the author implemented a suitable encoding and decoding solution representation method. This method was designed to ensure the suitability of the three proposed algorithms in solving the UPMSP problem effectively. According to the reported numerical results, the SOS-based methods achieved high-quality solutions within a reasonable time frame. The enhanced SOS algorithm and the hybridized SOS with SA algorithm demonstrated their effectiveness in solving the non-preemptive (UPMSP), showcasing improvements in solution quality and computational efficiency (Ezugwu, 2019).

### 2.2.3 Human-based optimization algorithms for UPMSP

Salimifard et al. (2020) tackled the problem of Parallel Machine Scheduling Problems (PMSPs) by formulating it as a bi-objective integer linear programming model. The model aimed to minimize two objectives: total tardiness and the number of waste. To address these issues, the study proposed a novel metaheuristic algorithm called Multi-Objective Volleyball Premier League (MOVPL), which evolved from the crowding distance concept used in NSGA-II. MOVPL was an extension of the Volleyball Premier League (VPL) method. The proposed approach was applied to ten test problems, and the results showed interesting performance. The MOVPL algorithm demonstrated its effectiveness in optimizing the bi-objective PMSPs by minimizing total tardiness and the number of waste. As future work, the study suggested enhancing MOVPL to handle additional constraints such as preemption, maintenance times, and uncertainty in the volume of jobs. These enhancements would further improve the algorithm's capability to address real-world PMSPs scenarios.

In their research, Rabiee et al. (2016) developed a biogeography-based optimization (BBO) approach to address the no-wait hybrid flow shop scheduling problem. This problem is characterized by realistic assumptions, including machine eligibility, sequence-dependent set-up times, and different ready times. The objective function of the study was focused on minimizing mean tardiness. To evaluate the impact of parameters on BBO, the researchers employed response surface methodology (RSM). They used mean relative percentage deviation (RPD), the standard deviation of RPD, best RPD, and worst RPD as evaluation metrics. The study demonstrated that BBO outperformed other existing solutions, both for small and large problem instances. The evaluation metrics consistently showed the superiority of the BBO approach. These findings highlight the effectiveness of BBO in tackling the no-wait hybrid flow shop scheduling problem. The study also suggested that further advancements could be made to the proposed solution when considering more complex scheduling assumptions. For instance, exploring the effects of sequence-dependent set-up times (SDSTs) and deterioration rates in processing times could enhance the performance of the BBO approach in handling intricate scheduling scenarios.

## 2.2.4 Trajectory-based optimization algorithms for UPMS

Haddad et al. (2015) applied AIV: a heuristic algorithm based on iterated local search and variable neighborhood descent for solving the UPMSP and the HIVP, which includes Path Relinking (PR) to generate a greedy initial solution and a partially greedy procedure to construct the initial solution, respectively. The approach was aimed at solving UPMSPST related problem. The study developed both AIV and HIVP from the Iterated Local Search (ILS) and Variable Neighborhood Descent (VND). And using benchmark test problems, statistical analysis revealed that AIV and HIVP attained an impressive performance compared with other approaches. Meanwhile, the study also showed that HIPV outperformed AIV even though they both performed better than the GA. The authors recommended that Mixed Integer Programming (MIP) model be incorporation into the AIV or HIVP to improving the performance of the method (Haddad, Cota, Souza, & Maculan, 2015).

Santos and Vilarinho (2010) proposed the utilization of simulated annealing (SA) to address the challenges posed by UPMSPs. The focus of their work was on UPMSPs with sequence-dependent setup, involving constraints such as equipment capacity, task precedence, lot sizing, and task delivery plan. The approach involved leveraging SA to manage the task's size in relation to available equipment, allocating larger widths to compatible equipment and distributing jobs to less utilized equipment. However, no comparison with other metaheuristic algorithms was conducted, although the study claimed to outperform previous mathematical programming approaches used for the same problem.

Anagnostopoulos and Rabadi (2002) present an innovative solution to the challenging scheduling problem of UPMS. The authors effectively introduce the problem of scheduling jobs on unrelated parallel machines with sequence-dependent setup times to minimize the makespan. They implement simulated annealing, a well-known metaheuristic algorithm, to minimize job completion time on unrelated parallel machines. The article provides a clear explanation of the simulated annealing algorithm and demonstrates its superior performance compared to existing methods. The inclusion of extensive experiments and statistical analysis enhances the credibility of the findings. While there is room for improvement in the presentation of results and discussion of practical implementation considerations, overall, this article contributes significantly to the field of parallel machine scheduling.

Silva et al. (2018) aimed to minimize the maximum completion time in the scheduling of UPMSPs with Sequence-Dependent Setup Times. The study compared the performance of five metaheuristic algorithms: Variable Neighborhood Search (VNS), Fix-and-Optimize (FO), Genetic Algorithm (GA), Relax-and-Fix (RF), and an exact solution method. Among the algorithms tested, VNS and FO exhibited more impressive results compared to GA, RF, and the exact solution method. VNS was particularly effective in finding better solutions and escaping local optima. In small problem instances, VNS and FO outperformed an improved version of GA (GA2). However, for larger problem instances,

GA2 outperformed VNS. In another study, Tozzo et al. (2018) investigated and compared the performance of GA and VNS in addressing UPMSPs with sequence machine-dependent setup time. The objective function of the problem was to minimize the makespan. The results indicated that VNS outperformed GA in terms of reducing the makespan. The authors recommended further research using a multiobjective function to assess the performance of both metaheuristic algorithms. Overall, these studies highlight the effectiveness of VNS and FO in minimizing the maximum completion time for UPMSPs with sequence-dependent setup times. VNS showed strong performance in finding optimal solutions and escaping local optima, while GA demonstrated effectiveness in larger problem instances. The comparisons provide insights for selecting appropriate metaheuristic algorithms based on problem characteristics and objectives.

Nogueira et al. (2014) tackled the problem of UPMSPs using the Greedy Randomized Adaptive Search Procedure (GRASP) metaheuristic algorithm. They employed a multiobjective approach that considered total earliness and tardiness penalties as performance measures. The method combined GRASP with a procedure based on the Path Relinking technique and Iterated Local Search (ILS) to find near-optimal solutions. Experimental results, obtained by applying the method to small, medium, and large instances, indicated that the hybridization of GRASP with Path Relinking (GRASP+PR) and GRASP with Iterated Local Search and Path Relinking (GRASP+ILS+PR) outperformed the standalone GRASP solution. The authors recommended exploring the use of the Variable Neighborhood Descent metaheuristic in the local search phase of ILS for future investigations (Nogueira, Arroyo, Villadiego, & Gonçalves, 2014).

In a similar vein, Gatica et al. (2013) conducted a study comparing the performance of GRASP, ILS, Simulated Annealing (SA), and Variable Neighborhood Search (VNS) in minimizing the Maximum Tardiness objective function in UPMSPs. The evaluation of the four single-solution algorithms revealed that SA outperformed the other three algorithms. However, the authors highlighted the need for future research to focus on hybridizing single-population metaheuristic algorithms with population-based algorithms such as Genetic Algorithms (GA) and Ant Colony Optimization (ACO) (Gatica, Esquivel, & Leguizamon, 2013).

Diana et al. (2014) presented a solution method for UPMSPs by hybridizing the Greedy Randomized Adaptive Search Procedure (GRASP) and Variable Neighborhood Descent (VND). The study incorporated a population re-selection operator to maintain the quality of the hybrid solution. GRASP was used for population generation, while VND was utilized as a somatic hypermutation operator to accelerate the convergence of GRASP. The authors argued that their proposed operator performed significantly better than a hybrid of GRASP and Genetic Algorithms (GA) (Diana, Filho, Souza, & Vitor, 2014).

In a similar work, Báez et al. (2019) addressed UPMSPs with dependent setup times by employing a hybrid of GRASP and Variable Neighborhood Search (VNS). Their objective was to minimize the total completion time by assigning jobs to machines and determining their processing order. The hybrid algorithm was applied to construct and improve optimal solutions. The study demonstrated that the hybrid approach outperformed other solutions, including exact methods. The authors suggested that future research should consider stochastic setup and processing times (Báez, Angel-Bello, Alvarez, & Melián-Batist, 2019).

Additionally, Yepes-Borrero et al. (2019) adapted the UPMSPs scheduling problem to include setup times and additional limited resources in the setups using the GRASP metaheuristic algorithm. The proposed GRASP approach involved two strategies: one that disregarded information about other resources during the construction phase and another that considered this information. The results showed that the solution method effectively addressed the problem. The authors recommended further exploration of the study in the context of scheduling problems such as the flowshop.

Lee et al. (2013) employed a Tabu Search (TS) heuristic algorithm with various neighborhood generation methods to address the UPMSP with sequence-dependent setup times. Their objective was to minimize the total tardiness. Experimental evaluations were conducted comparing TS with Simulated Annealing (SA) and Iterated Greedy algorithms. The comparative analysis demonstrated that TS outperformed the other algorithms by producing optimal solutions for both large and small problem instances, with over 50% of the solutions being optimal. However, the study also revealed that while TS provided a greater number of optimal solutions quickly, the quality of its solutions was not as high as those from greedy search algorithms. The authors suggested that future advancements could involve using advanced neighborhood generation methods with TS and applying them to stochastic problems with non-zero ready times (Lee, Yu, & Lee, 2013).

In another study, Shahvari and Logendran (2017) improved the TS algorithm to solve the UPMSP with sequence- and machine-dependent batch scheduling. Their approach introduced enhancements such as multi-level diversification, multi-tabu structure, and the use of lemmas to eliminate ineffective neighborhoods during the search process. The problem was formulated as a Mixed Integer Linear Programming (MILP) model and solved using the improved TS algorithm. The objective was to minimize a multi-objective function comprising total weighted completion time and total weighted tardiness. The results demonstrated that the improved TS algorithm successfully reduced the objective function value by 37%, and the incorporation of lemmas further improved it up to 40% within computational time constraints. The study proposed future research directions to explore the performance of a hybrid TS-based algorithm in combination with other metaheuristics (Shahvari & Logendran, 2017).

Similarly, Chen (2012) investigated the performance of a hybrid approach combining Iterated Local Search (ILS), TS, and Variable Neighborhood Descent (VND) to solve the UPMSP with sequence-dependent setup times and unequal ready times. The study integrated TS and VND with an Iterated Hybrid Metaheuristic (IHM) algorithm, referred to as ILS, to address the problem. Performance evaluations were conducted using processing times, ready times, and due-date tightness as evaluation criteria. The results showed that the proposed hybrid solution outperformed the individual metaheuristic algorithms when considering the specified evaluation criteria (Chen, 2012).

## 2.2.5 Hybrid based optimization algorithms for UPMS

Hybrid algorithms leverage the strengths of multiple optimization algorithms, leading to an enhanced optimization capability while minimizing computational complexities (Ting et al., 2015). This category of algorithms has garnered extensive research attention, resulting in a wide range of articles and variants. By combining the features of different algorithms, hybrid approaches offer promising avenues for improving optimization techniques and addressing complex problems effectively.

Jouhari, Lei, Abd, Ewee, and Farouk (2019) proposed a hybrid algorithm combining simulated annealing (SA) and the Sine Cosine Algorithm (SCA) to address UPMSPs and minimize the makespan. The hybrid algorithm, named SASCA, employed SCA as a local search method to enhance the performance and convergence of SA in obtaining efficient solutions. Numerical results indicated that SASCA demonstrated favorable performance in both small and large problem instances, outperforming other metaheuristic algorithms. In another related work, A hybridization of ACO, Simulated Annealing (SA), and Variable Neighborhood Search (VNS) was proposed to tackle UPMSPs with parallel machines and sequence-dependent setup times, aiming to reduce makespan. ACO and SA were utilized for solution evolution, while VNS aimed to improve the population. The study demonstrated that the hybridized algorithm produced high-quality solutions even for larger problem instances. Comparative analysis revealed that ACO, SA, and VNS outperformed hybrids of ACO and VNS, as well as SA and VNS. The authors suggested that the proposed hybrid algorithm should be extended to address UPMSPs with ready times for each job's start time as future work (Behnamian, Zandieh, & S.M.T., 2009). Zabihzadeh and Rezaeian (2015) addressed UPMSPs in a robot job environment using ACO with double pheromone and GA. The problem was formulated as a mixed-integer linear programming model, considering two objective functions: makespan, aiming to minimize the sequence of processing parts and robots' movements, and finding the closest number to the optimal number of robots. The study revealed that GA performed well in determining near-optimal numbers of robots compared to the results obtained from ACO. The authors recommended the inclusion of due dates and the consideration of objective functions such as maximum lateness, total tardiness, and the number of tardy jobs in future models (Zabihzadeh & Rezaeian, 2015).

Abbaszadeh, Asadi-Gangraj, and Emami (2020) tackled the Flexible Flow Shop (FFS) scheduling problem with UPMSPs and a renewable resource shared among the stages. They employed a combination of Simulated Annealing (SA) and Particle Swarm Optimization (PSO) algorithms, referred to as SA-PSO, to solve the problem. The study began by developing a Mixed-Integer Linear Programming (MILP) model to minimize the maximum completion time (makespan). Subsequently, PSO and the hybrid SA-PSO were employed to solve the model. The results demonstrated that the SA-PSO hybrid performed well, particularly for large-sized problems, outperforming PSO alone. The authors recommended further research into the performance of hybrid metaheuristic algorithms such as Genetic Algorithm (GA), Variable Neighborhood Search (VNS), and Tabu Search (TS). In a similar vein, Lin (2013) applied PSO to solve a scheduling problem involving UPMSPs where $n$ jobs were scheduled on $m$ UPMSPs, considering the presence of release dates. The objective function aimed to minimize the makespan and subsequently utilize the proposed method to maximize the utilization of machines. The results obtained in the study demonstrated that PSO yielded impressive results when compared to other metaheuristic algorithms. The authors suggested that the proposed solution method could be extended to address multi-objective parallel machine scheduling problems.

Mir and Rezaeian (2015) proposed a hybrid metaheuristic algorithm combining Particle Swarm Optimization (PSO) and Genetic Algorithm (GA) to solve UPMSPs with past-sequence-dependent setup times, release dates, deteriorating jobs, and learning effects. The study further enhanced the hybrid system by incorporating the Taguchi method to optimize and select the optimal parameters. Through experimentation, the hybrid metaheuristic algorithms were compared with and without local search. The results indicated that the latter algorithms were suitable for small problem instances, while the hybrid of PSO and GA outperformed them for large problem instances. The study also recommended exploring other metaheuristic algorithms and applying their solutions to multi-objective models, indicating potential avenues for future research. In a related work, Torabi and colleagues (2013) proposed the use of PSO with multi-objective criteria, known as Multi-Objective Particle Swarm Optimization (MOPSO), to solve a multi-objective model formulated as a UPMSPs problem. The objective functions considered were total weighted flow time, total weighted tardiness, and total machine load variation. The performance analysis of MOPSO compared to Conventional Multi-Objective Particle Swarm Optimization (CMOPSO) using randomly generated test problems demonstrated that MOPSO outperformed CMOPSO. The study highlighted the effectiveness of MOPSO in finding a good approximation of the Pareto frontier.

Kerkhove and Vanhoucke (2014) addressed the problem of the UPMSP in the context of a Belgian producer of knitted fabrics. The objective was to assign N jobs with release dates to M machines with due dates while considering changeover times and sequence-dependent setup times to minimize the weighted combination of job lateness and tardiness. To tackle this problem, the authors employed a hybrid metaheuristic algorithm combining Simulated Annealing (SA) and Genetic Algorithm (GA). The SA-GA hybrid algorithm proved effective in solving real-scale scheduling problems, with instances of up to 750 jobs, 75 machines, and 10 production locations, within a reasonable computational time. To mitigate the impact of changeover interference, the algorithm was complemented with heuristic dispatching rules that prioritized shorter changeover times. The study demonstrated a performance increase of 23% when up to 12 machines were serviced due to the support provided by the heuristic rules.

Jolai et al. (2012) tackled the problem of a no-wait flexible flow shop manufacturing system with sequence-dependent setup times. Their objective was to minimize the maximum completion time. To address the problem, they improved the Simulated Annealing (SA) and Imperialist Competitive Algorithm (ICA) versions. Population-Based Simulated Annealing (PBSA) and Adapted Imperialist Competitive Algorithm (AICA) were used as hybrid algorithms. The authors employed the Taguchi method for parameter optimization. The study demonstrated that the hybrid algorithm outperformed the individual metaheuristic algorithms (Jolai, Rabiee, & Asefi, 2012).

In a related study, Garavito-Hernández et al. (2019) explored a hybrid approach combining ICA and GA to solve UPMSPs in a flow shop (HFS) scheduling context with sequence and machine-dependent setup times. Their solution was compared with mixed-integer programming models using an exact method. The results showed that their hybrid solution performed equally or better in terms of providing solutions (Garavito-Hernández, Peña-Tibaduiza, Perez-Figueredo, & Moratto-Chimenty, 2019). The study also recommended enhancing meta-heuristic algorithms by incorporating local search strategies for solution improvement.

Rahmanidoust et al. (2017) conducted a study to investigate the performances of Harmony Search (HS), Imperialist Competitive Algorithm (ICA), and a hybridization of PBSA and ICA (ICA+PBSA) when applied to solve the problem of no-wait hybrid flow shop scheduling in relation to UPMSPs. The objective of the study was to minimize the mean tardiness while addressing four specific challenges: no-wait operations, separate setup time of each job from its processing time, in coordination with job arrival time, and inconsistent machine availability. By using the Taguchi approach for parameter definition and random test problems, the study demonstrated that HS outperformed the other algorithms in terms of performance. As future work, the authors suggested investigating novel meta-heuristics such as the FA and graph coloring-based algorithms for solving the problem. Additionally, the authors proposed exploring models characterized by emergency maintenance, learning effect, and deterioration, as they may yield interesting performance outcomes. Furthermore, Rahmanidoust et al. (2017) explored various meta-heuristic algorithms, including HS, ICA, and ICA+PBSA, to solve the problem of no-wait hybrid flow shop scheduling in UPMSPs. The study highlighted the superior performance of HS and recommended further research into alternative meta-heuristic approaches and the incorporation of additional problem characteristics.

In their study, Sadati et al. tackled an industrial problem concerning unmanned production maintenance processes (UPMPs) by applying a hybrid multi-objective teaching-learning based optimization (HMOTLBO) approach. To effectively address the problem, they formulated it as a mixed-integer linear programming (MILP) model. By utilizing the ε-constraint method, the researchers solved the multi-objective model using HMOTLBO, specifically focusing on small-sized problems. To assess the performance of the proposed HMOTLBO approach, it was compared with a non-dominated sorting genetic algorithm (NSGA-II). The results indicated that the hybrid method exhibited favorable performance. However, the study acknowledged the potential for further improvements and suggested that future research endeavors should explore the incorporation of additional constraints. These constraints could include factors such as pre-emption, precedence constraints, and machine failures (Sadati, Tavakkoli-Moghaddam, Naderic, & Mohammadi, 2017).

### 2.2.6 Machine Learning based optimization algorithms for UPMS

Cheng et al. (2020) addressed the problem of UPMSPs in the forging industry, specifically focusing on uncertain machine-dependent and job sequence-dependent setup times (MDJSDSTs). They proposed a metaheuristic algorithm based on RandomForest to minimize the makespan. The aim was to improve the estimation of setup times for large instances. Experimental results demonstrated a significant reduction in the error percentage for setup time estimation using their method. The authors suggested that future studies could explore the problem's performance by incorporating simulation-based optimization approaches and considering factors such as processing time and waiting time (Cheng, Pourhejazy, Ying, Li, & Chang, 2020).

In their study, Park et al. (2000) propose a novel approach for scheduling jobs with sequence-dependent setup times on parallel machines. The authors combine a neural network with heuristic rules to effectively handle the scheduling problem. Notably, they utilize a neural network to calculate the priority index of each job, which plays a crucial role in the scheduling process. The integration of a neural network in the scheduling algorithm brings forth the advantages of machine learning techniques, allowing for more sophisticated and data-driven decision-making. By training the neural network on relevant job characteristics and historical data, the authors were able to derive accurate priority indices that reflect the importance and urgency of each job. In the comprehensive review conducted by Đurasević and Jakobović (2023), multiple investigations into the current cutting-edge research on scheduling tasks for parallel machines were undertaken and thoroughly deliberated upon. Henceforth, readers with an interest in this subject are directed to the previously mentioned survey paper for further information.

Figures 1, 2, and 3 provide an overview of the utilization of various optimization algorithms in addressing the problem discussed in this study over the past two decades. These figures depict the rate at which these algorithms have been applied and utilized. Additionally, an illustration showcasing the hybridization of these metaheuristic algorithms is presented.

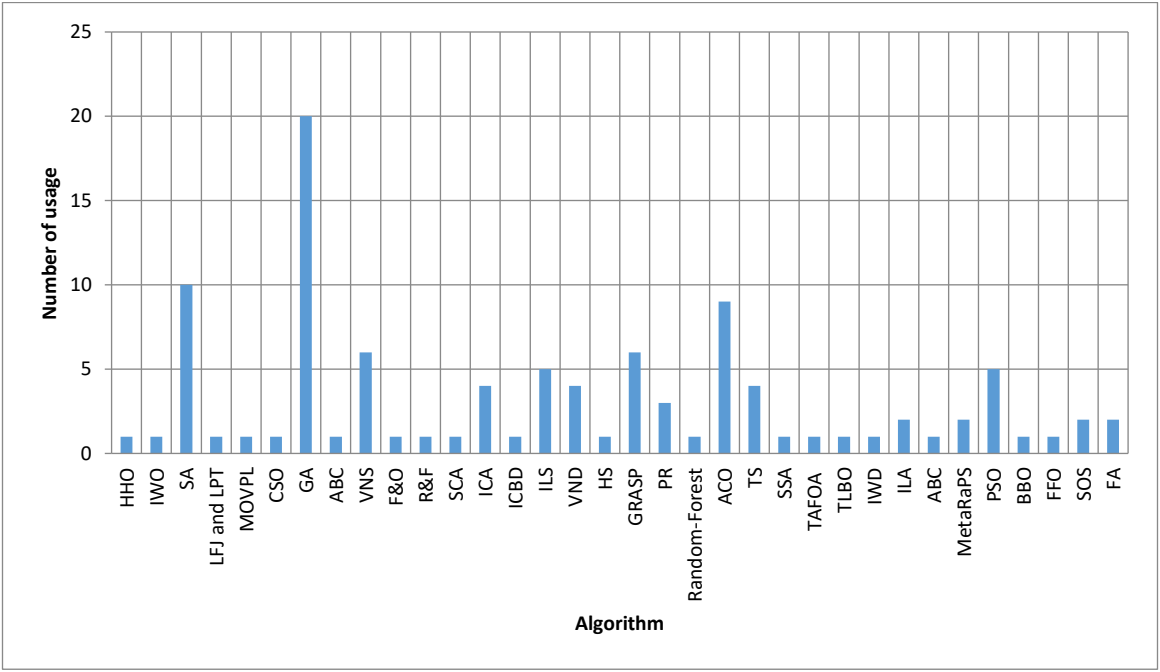

Fig. 1: A distribution of the choice of metaheuristic algorithms for applicability to the problem of UPMSPs

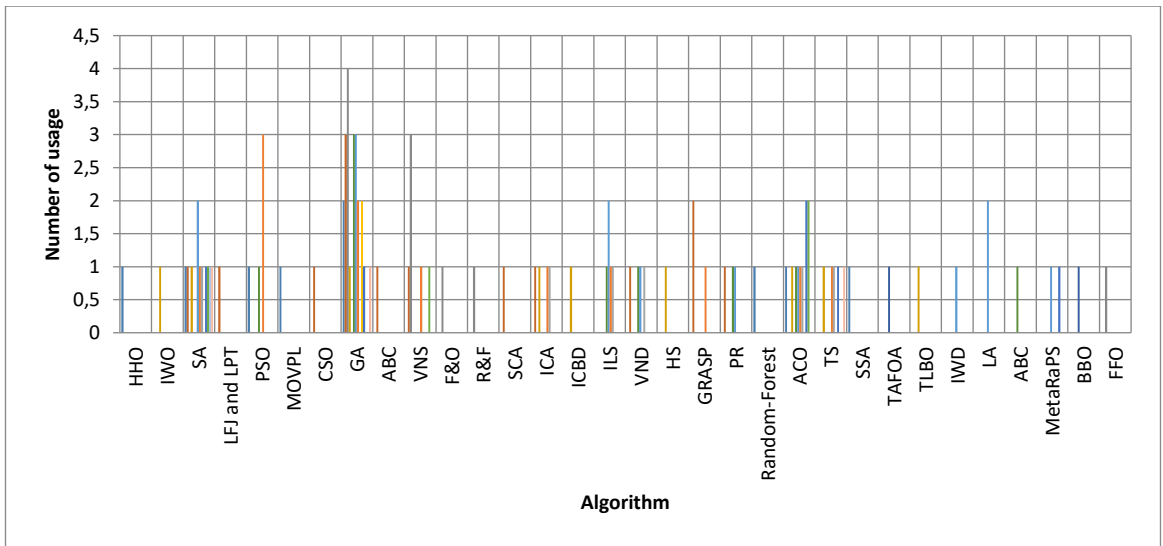

Fig. 2: A yearly based report on how each metaheuristic algorithms are used to optimize the problem of UPMSPs

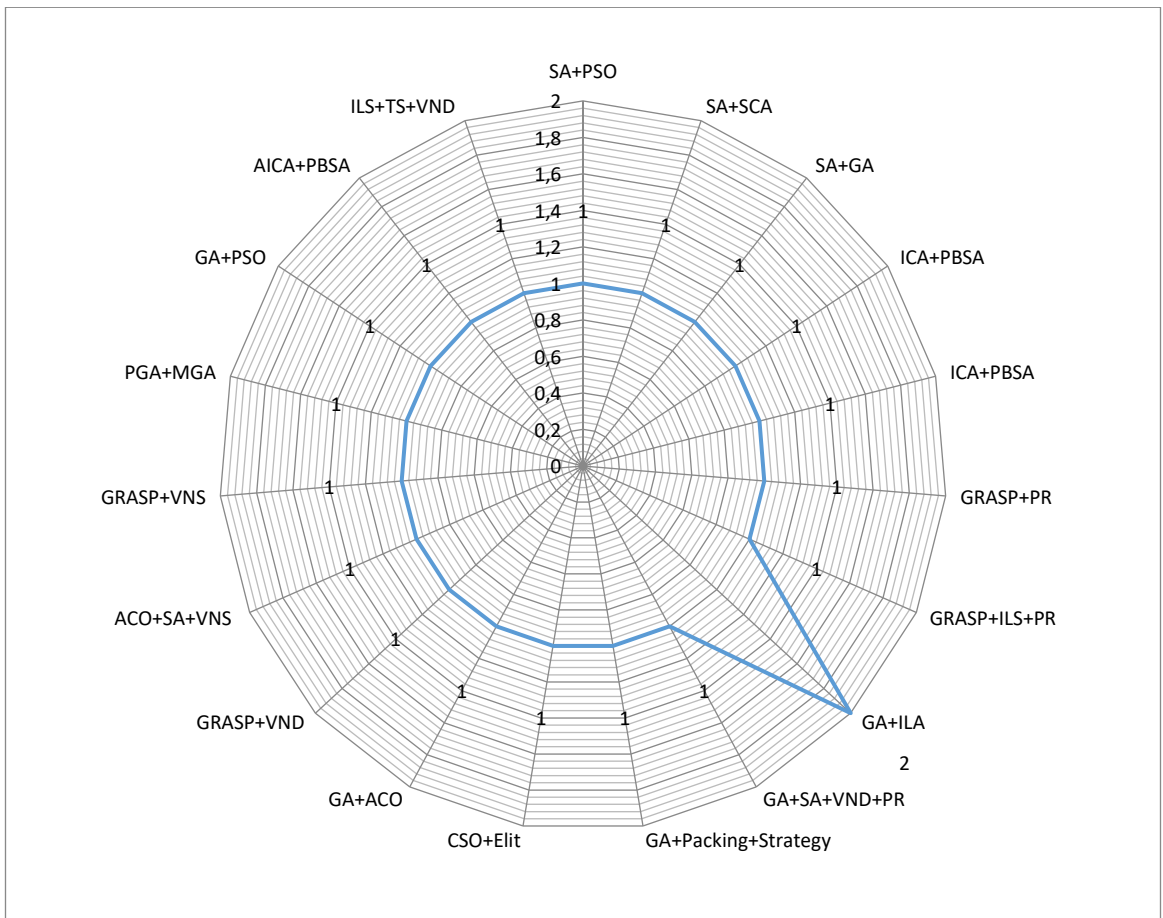

Fig. 3: Illustration of hybrid metaheuristic algorithms that have been employed to solve the UPMSPs

## 3. Problem Formulation and Mathematical Model

In this section, we discuss the mixed integer programming (MIP) model, a commonly used mathematical formulation for representing parallel machine scheduling problems. The MIP model is effective in defining the objective function and constraints of the problem accurately. In this study, we focus on a variant of the UPMSP that minimizes the total completion time, taking into consideration the setup times of individual machines. Moreover, the MIP solvers are advantageous in finding optimal solutions, but they often require significant computational time. Therefore, they are more suitable for solving small-sized problem instances. In this paper, we adopt and enhance the MIP formulation presented in previous studies for minimizing the makespan of the UPMSP (Anagnostopoulos and Rabadi, 2002; Helal, Rabadi, and Al-Salem, 2006; Ezugwu and Akutsah, 2018; Ezugwu, 2019; 2022). We believe that our MIP model provides a clear theoretical foundation for the variant of the parallel machine problem addressed in this research. The variant of the UPMS problem described and formulated here is of the form $P_m|S_{i,j,k}|C_{max}$. The mathematical model formulation for the problem at hand is presented as follows:

**Sets and Indices**
$N$: Number of jobs to be processed, $N = \{1,2, \dots, n\}$
$M$: Number of machines, $M = \{1,2, \dots, m\}$
$i, j$: Job indices, where $i, j \in N$
$k$: Machine indices, where $k \in M$

**Parameters**
$P_{j,k}$: Processing time of job $j$ on machine $k$
$S_{i,j,k}$: Sequence-dependent setup time if job $j$ is scheduled directly after job $i$ on machine $k$
$V$: a large positive number

**Decision variables**
$C_{max}$: Maximum completion time (or makespan)
$C_j$: Completion time of job $j$
$AP_{i,j,k}$: Adjusted processing time matrix of job $j$ when it is processed immediately after job $i$ on machine $k$
$x_{i,j,k}$: 1 if job $j$ is scheduled directly after job $i$ on machine $k$ and 0 otherwise
$S_{0,j,k}$: Setup time if job $j$ is scheduled to go first on machine $k$
$x_{0,j,k}$: 1 if job $j$ is scheduled first on machine $k$ and 0 otherwise
$x_{i,0,k}$: 1 if job $j$ is scheduled last on machine $k$ and 0 otherwise

**Model formulation**

$$Min\ C_{max} \tag{1}$$

Subject to

$$\sum_{k=1}^{m}\sum_{\substack{i=0\\i\neq j}}^{n} x_{i,j,k} = 1, \quad \forall j = 1, \dots, n \tag{2}$$

$$\sum_{j=1}^{n} x_{0,j,k} = 1, \quad \forall k = 1, \dots, m \tag{3}$$

$$\sum_{\substack{i=0\\i\neq h}}^{n} x_{i,j,h} - \sum_{\substack{j=0\\j\neq h}}^{n} x_{h,j,k} = 0, \quad \forall h = 1, \dots, n, \forall k = 1, \dots, m \tag{4}$$

$$C_j - \left[ C_i + \sum_{k=1}^{m} x_{i,j,k}\left(S_{i,j,k} + P_{j,k}\right) + V\left(\sum_{k=1}^{m} x_{i,j,k} - 1\right) \right] \geq 0, \quad \forall i = 0, \dots, n, \forall j = 1, \dots, n \tag{5}$$

$$C_0 = 0 \tag{6}$$

$$x_{i,j,k} \in \{0,1\}, \forall i = 0, \dots, n, \forall j = 0, \dots, n, \forall k = 1, \dots, m \tag{7}$$

**Constraints descriptions:**
Equation 1 represents the objective function of the problem, aiming to minimize the makespan. Equation 2 ensures that each job is scheduled only once and processed by a single machine. Equation 3 guarantees that only one job can be scheduled as the first job and not more than once. Equation 4 ensures that each job must have one preceding job and one succeeding job. Equation 5 calculates the completion times of the jobs and ensures that no job can both precede and succeed the same job. Equation 6 sets the completion time for the initial dummy job to zero. Equation 7 specifies that the decision variable $x_{i,j,k}$ is binary across all domains.

Note that the processing times are given in a matrix $[P]$ of size $m \times n$, and the setup times are given in a set of $m$ matrices $[S]_k$, each of size $n \times n$. The $[AP]_k$ matrix for each machine is defined such that the following constraints hold: $\forall k = 1, \dots, m, \forall j = 1, \dots, n$: $AP_{i,j,k} = S_{i,j,k} + P_{j,k}, \forall i = 1, \dots, n$. Therefore, instead of considering $[P]$ and $[S]_k$, we introduce the concept of Adjusted Processing Time Matrix $[AP]_k$ into Equation (5) mentioned earlier for each machine $k$ such that:

$$C_j - \left[ C_i + \sum_{k=1}^{m} x_{i,j,k} AP_{i,j,k} + V\left(\sum_{k=1}^{m} x_{i,j,k} - 1\right) \right] \geq 0, \quad \forall i = 0, \dots, n, \forall j = 1, \dots, n \tag{8}$$

The makespan, which is the maximum completion time required to process jobs $i$ and $j$ on machine $k$, is therefore given by:

$$C_{max} = max_{1 \leq j \leq n}\{C_j\} = \max_{\substack{k=1,\dots,m\\i=1,\dots,n}} \sum_{j=1}^{n} [AP_{i,j,k}] \tag{9}$$

This equation is derived from the fact that the makespan is the maximum time required to complete all jobs, and the time required to complete a job is the sum of its processing time and setup time. The adjusted equation (8) takes into account the fact that the setup time for a job may vary depending on the machine on which it is processed. The adjusted equation (8) has been used in previous studies to solve the UPMSP (Ezugwu and Akutsah, 2019; Ezugwu, 2019; Ezugwu, 2022). In these studies, the adjusted equation was shown to be effective in finding optimal solutions to the UPMSP.

The objective function for the UPMSP is to minimize the maximum completion time, as expressed in equation (10). This can be written as follows, as discussed in previous works (Ezugwu and Akutsah, 2019; Ezugwu, 2019; Ezugwu, 2022):

$$Minimize \quad C_{max} = \max_{\substack{k=1,\ldots,m \\ i=1,\ldots,n}} \sum_{j=1}^{n} \left[ AP_{i,j,k} \right]$$

(10)

## 4. Application of Metaheuristics for UPMS Problems

In this section, we provide a brief overview of nine well-known metaheuristic algorithms selected for solving the specific UPMS problem discussed in this paper. The choice of these nine algorithms was influenced by their individual strengths and successful application history in tackling various challenging variants of the UPMS problem. Moreover, their superior performances, as reported in related optimization literature, also supported their selection (Ezugwu et al., 2020; Vallada, Villa, and Fanjul-Peyro, 2019; Abdeljaoued, Saadani, and Bahroun, 2020; Moser et al., 2022). In order to ensure consistency and follow of thought, some concepts from previous related published work were intentionally adapted in this paper. Specifically, the following concepts were replicated and represented:

- Adjusted Processing Times Matrix (Section 4.1)
- Solution Representations (Section 4.2)
- Initialization or Initial Solutions (Section 4.3)
- Mutation-based Local Search Improvement Schemes (Section 4.4)

These concepts were adapted from the work of Ezugwu and Akutsah (2018), Ezugwu (2019), and Ezugwu (2022), and discussed in more detail in this paper.

The nine representative algorithms chosen for this study are as follows: FA, ACO, GA, IWO, ABC, BA, DE, PSO, and TLBO. For each of these algorithms, we adopted the standard algorithmic design structures as described in the original literature with minor modifications to adapt them for solving the specific UPMS problem. These modifications include the implementation of solution representation, initial solution generation, fitness value evaluation, and mutation-based local search improvement schemes.

It is important to note that we retained the basic algorithmic structures of each representative algorithm from the authors' original works to ensure an unbiased performance comparative analysis. Due to page limitations and to avoid redundant presentations of existing concepts, interested readers are referred to explore the original representations of these representative algorithms in their respective literature sources.

These sources include Holland (1992) for GA, Kennedy and Eberhart (1995) for PSO, Dorigo, Birattari, and Stutzle (2006) for ACO and ABC, Karimkashi and Kishk (2010) for IWO, Price (2013) for DE, and Rao, Savsani, and Vakharia (2011) for TLBO. FA was developed by Yang and He (2013).

First, we adapted a generalized algorithmic representation of a typical population-based metaheuristic optimization algorithm design. This representation was used to denote the optimization perspectives for all the representative algorithms selected to handle the UPMS problems. The standard algorithmic design structure is presented in Algorithm Listing 1.

**Algorithm 1**: Generalized algorithmic representation of population based metaheuristics

1. *Initialize the population of solutions randomly or using some specific initialization strategy.*
2. *Evaluate the fitness or objective function value of each solution in the population.*
3. *Set the initial temperature or exploration parameter (if applicable) for the algorithm.*
4. *Repeat until a termination condition is met (e.g., maximum iterations, convergence criteria):*
   a. *Perform local search or exploitation to improve the solutions in the population.*
   b. *Perform global search or exploration to explore new areas of the search space.*
   c. *Update the fitness values of the solutions.*
   d. *Determine the best solution found so far and update if necessary.*
   e. *Adapt or update any algorithm-specific parameters (e.g., cooling schedule, mutation rate).*
5. *Return the best solution obtained during the optimization process.*

This generalized representation encapsulates the iterative nature of metaheuristic algorithms, which integrate local search and global search strategies to effectively explore and exploit the search space. However, it is important to note that the specific implementation of each step may vary depending on the particular metaheuristic algorithm employed and the specific problem under consideration.

Similarly, the generalized algorithmic representation presented in algorithm listing 2 provides a framework for the population-based metaheuristic optimization used to solve the UPMS problems with some added evolutionary features. Each representative evolutionary algorithm such as the GA follows this structure, but with algorithm-specific modifications and operators tailored to the algorithm's approach. By employing this standardized structure, we can compare and analyze the performance of the different algorithms based on their ability to find optimal schedule configurations with minimal makespan.

**Algorithm 2**: Algorithmic design for population-based metaheuristic optimization with evolutionary adaptation features for UPMS problems

*Input:*
- *Problem-specific details (UPMSP problem formulation, constraints, etc.)*
- *Population size (N)*
- *Maximum number of iterations (max_iter)*

*Output:*
- *Best solution found (optimal schedule configuration)*

- *Objective function value of the best solution (minimum makespan)*

*Procedure:*
1. *Initialize a population of N candidate solutions randomly within the problem search space.*
2. *Evaluate the fitness value of each candidate solution using the objective function (maximum completion time).*
3. *Set the best solution found as the initial solution with the lowest fitness value.*
4. *Iterate until reaching the maximum number of iterations or a stopping criterion is met:*
   a. *Perform selection operation to choose candidate solutions for reproduction and further exploration.*
   b. *Apply crossover and mutation operators to generate new candidate solutions.*
   c. *Evaluate the fitness value of each new candidate solution.*
   d. *Update the best solution found if a new solution with a lower fitness value is discovered.*
5. *Return the best solution found and its corresponding objective function value.*

Note: The specific implementation of selection, crossover, and mutation operators may vary depending on the chosen metaheuristic algorithm and problem characteristics.

### 4.1. Solution representation

The nine representative metaheuristic optimization algorithms employed in this study are all population-based optimization methods. The optimization process begins by initializing a population of candidate solutions, denoted by the parameter $X$, according to Equation 11. The candidate solutions are randomly generated within the given problem landscape, bounded by the upper and lower limits. Throughout the iterative processes of the representative algorithms, the best-recorded solution is continually updated and considered as the optimal solution (Ezugwu, 2022).

$$X = \begin{bmatrix} X_1 \\ \ldots \\ X_i \\ \vdots \\ X_{N-1} \\ X_N \end{bmatrix} = \begin{bmatrix} x_{1,1} & \ldots & x_{1,j} & x_{1,D-1} & x_{1,D} \\ x_{2,1} & \ldots & x_{2,j} & \ldots & x_{2,D} \\ \ldots & \ldots & x_{i,j} & \ldots & \ldots \\ \vdots & \vdots & \vdots & \vdots & \vdots \\ x_{N-1,1} & \ldots & x_{N-1,j} & \ldots & x_{N-1,D} \\ x_{N,1} & \ldots & x_{N,j} & x_{N,D-1} & x_{N,D} \end{bmatrix} \quad (11)$$

where,
$$x_{i,j} = rand \times (UB - LB) + LB, i = 1,2,\ldots,N, j = 1,2,\ldots,D \quad (12)$$

The set of all possible solutions, denoted by $X$, is generated randomly based on the model equation given in equation 12. The decision variable for the ith candidate solution is represented by $X_i$, and it takes values within the search space $X$. The parameter $N$ represents the population size of candidate solutions, and $D$ indicates the dimension of the problem search space. The parameters LB and UB correspond to the lower and upper bounds, respectively.

Based on the solution initialization procedure described above, the individual representative algorithms are designed to effectively solve the UPMSP within the solution search space $X$. This search space consists of all possible schedule configurations for $n$ jobs that can be scheduled on $m$ machines. To adapt the selected algorithms for handling the specific scheduling problem of the form $P_m|S_{i,j,k}|C_{max}$, a local search improvement algorithm is incorporated to enhance their solution search capabilities. The details of the local search approach can be found in algorithm listing 2.

$$X_i = \begin{bmatrix} X_1 \\ \dots \\ X_i \\ \vdots \\ X_{m-1} \\ X_m \end{bmatrix} = \begin{bmatrix} x_{1,1} & \dots & x_{1,j} & x_{1,n-1} & x_{1,n} & 0 & 0 & 0 & 0 & 0 & 0 & 0 & 0 & 0 & 0 & 0 \\ x_{2,1} & \dots & x_{2,j} & \dots & x_{2,n} & 0 & 0 & 0 & 0 & 0 & 0 & 0 & 0 & 0 & 0 & 0 \\ \dots & \dots & x_{i,j} & \dots & \dots & 0 & 0 & 0 & 0 & 0 & 0 & 0 & 0 & 0 & 0 & 0 \\ \vdots & \vdots & \vdots & \vdots & \vdots & 0 & 0 & 0 & 0 & 0 & 0 & 0 & 0 & 0 & 0 & 0 \\ x_{m-1,1} & \dots & x_{m-1,j} & \dots & x_{m-1,n} & 0 & 0 & 0 & 0 & 0 & 0 & 0 & 0 & 0 & 0 & 0 \\ x_{m,1} & \dots & x_{m,j} & x_{m,n-1} & x_{m,n} & 0 & 0 & 0 & 0 & 0 & 0 & 0 & 0 & 0 & 0 & 0 \end{bmatrix}$$
(13)

In the first row of the matrix representation, the zero after job $x_{1,n}$ indicates that job $x_{1,n}$ is the last job to be processed by machine 1. This convention applies to the remaining rows as well. Specifically, the zeros after jobs $x_{2,n}$, $x_{k,n}$ and $x_{m,n}$ indicate that these jobs are the last ones to be processed by machines 2, $k$ and $m$, respectively.

The solution representation encoding, as described in (Ezugwu, 2022), is utilized for machine assignment and scheduling sequencing. In terms of machine assignment, a two-stage solution representation is employed. The first stage involves assigning n jobs to m unrelated parallel machines, aiming to minimize the maximum completion time ($C_{max}$) among all machines. This assignment is represented as an integer vector, denoted as $x_1$, with a dimension equal to the number of jobs (Ezugwu, 2019; Ezugwu, 2022).

For example, considering 16 jobs ($n = 16$) and 4 machines ($m = 4$), the vector $x_1 = [4, 1, 4, 4, 4, 1, 3, 3, 3, 3, 2, 2, 2, 2, 1, 1]$ implies that the first machine, $m_1$, will be assigned jobs 2, 6, 15, and 16; the second machine, $m_2$, will be assigned jobs 11, 12, 13, and 14; the third machine, $m_3$, will be assigned jobs 7, 8, 9, and 10; and the fourth machine, $m_4$, will be assigned jobs 1, 3, 4, and 5.

The second stage involves the job sequencing operation, represented in matrix form with the same dimensions as the machine assignment vector. This job sequencing, denoted as $x_2$, can be represented as an $m \times n$ matrix, illustrating the sequencing operations on each machine. Equation 14 (Ezugwu and Akutsah, 2018; Ezugwu, 2022) provides a detailed illustration of this matrix representation for job sequencing operations.

$$x_2 = \begin{bmatrix} 6 & 15 & 16 & 2 & 0 & 0 & 0 & 0 & 0 & 0 & 0 & 0 & 0 & 0 & 0 & 0 \\ 12 & 11 & 13 & 14 & 0 & 0 & 0 & 0 & 0 & 0 & 0 & 0 & 0 & 0 & 0 & 0 \\ 8 & 9 & 10 & 7 & 0 & 0 & 0 & 0 & 0 & 0 & 0 & 0 & 0 & 0 & 0 & 0 \\ 3 & 4 & 5 & 1 & 0 & 0 & 0 & 0 & 0 & 0 & 0 & 0 & 0 & 0 & 0 & 0 \end{bmatrix} \quad (14)$$

Building upon the example illustrations discussed in previous studies (Ezugwu and Akutsah, 2018; Ezugwu, 2019), the variables $x_1$ and $x_2$ describe the sequence of operations for each machine. For instance, in machine $m_1$, the sequence operation consists of job 6, job 15, job 16, and job 2. The same sequence description applies to machines $m_2$ and $m_3$.

The presence of zeros after certain jobs, such as job 2, job 14, job 7, and job 1, indicates that these jobs are the last to be processed by their respective machines ($m_1$ for job 2, $m_2$ for job 14, $m_3$ for job 7, and $m_4$ for job 1).

It is important to note that the assignment and sequencing mechanisms are based on the individual solutions or population variations. For instance, in the case of the FA, a notable global optimization metaheuristic, the variation in the firefly's light intensity is taken into account for the assignment and sequencing mechanisms during the algorithm's implementation for the scheduling sequence operation test problem. The light intensities are adjusted based on the quality of solutions in each generation.

In the context of the FA, the light intensity, denoted as $I$, representing the solution $x$, is proportional to the value of the fitness function, $I(x) \propto f(x)$. According to (Yang, 2010), the light intensity, $I$, varies according to the following equation (Ezugwu, 2022):
$$I(r) = I_0 e^{-\gamma r^2} \tag{15}$$

Here, $I_0$ represents the light intensity of the source, and $\gamma$ represents the light absorption coefficient. This equation describes the variation of light intensity, $I$, with respect to the distance parameter, $r$.

**4.2. Why Firefly algorithm?**

The FA is regarded as the superior and favored optimization approach for the demonstrated proof of concept related to employing metaheuristics for sustainable scheduling. This preference stems from the algorithm's robust capabilities in effectively addressing challenging and intricate optimization problems, particularly those characterized by extremeness and complexity. Further more, several notable advantages of using the FA for solving complex real-world optimization problems and other related scheduling of unrelated parallel machines tasks include (Yang and Slowik 2020; Zhang et al. 2016; Arora and Singh, 2013; Đurasević and Jakobović, 2023):

- Global optimization: The FA is capable of exploring a wide solution space, making it effective in finding globally optimal or near-optimal solutions for complex scheduling problems involving unrelated parallel machines.
- Diversity preservation: The algorithm's exploration mechanism helps maintain diverse solutions across the population of fireflies, preventing premature convergence and aiding in escaping local optima.
- Adaptability: FA can be easily adapted to different scheduling objectives and constraints. This flexibility allows it to handle a variety of scheduling scenarios, making it applicable to various real-world industrial problems.
- Parallelism: The algorithm inherently operates in a parallel manner, which aligns well with the nature of scheduling for parallel machines. This can lead to faster convergence and reduced computation time.
- Heuristic nature: FA is a heuristic optimization technique, which means it doesn't require explicit gradient information and can work well even for non-convex and nonlinear objective functions, commonly encountered in scheduling problems.

- Ease of implementation: The algorithm's simple rules for firefly movement and attractiveness enable relatively straightforward implementation and experimentation.
- Exploration-exploitation balance: The FA employs a balance between exploration and exploitation, allowing it to explore the search space efficiently while refining solutions as the algorithm progresses.
- Robustness: The algorithm's ability to escape local optima and adapt to various problem instances contributes to its robustness in handling scheduling challenges, even in dynamic or uncertain environments.
- Non-derivative-based: The FA doesn't rely on derivative information, making it suitable for problems where gradients are hard to obtain or compute.
- Effective for complex problems: Scheduling unrelated parallel machines can involve intricate relationships between tasks, machines, and objectives. The FA ability to handle complex, multi-dimensional optimization spaces makes it a valuable tool in this context.

It's worth noting that while the FA offers several advantages, its performance can still be influenced by parameter tuning, problem characteristics, and the specific implementation details. Therefore, careful experimentation and analysis are essential to harness its full potential for scheduling unrelated parallel machines effectively. Algorithm Listing 3 introduces the FA optimization representation designed for solving the Unrelated Parallel Machine Scheduling Problem (UPMSP). It's worth mentioning that all hybrid versions of the FA scheduling approach are implemented by making minor adjustments to the core FA base algorithm, illustrated in Algorithm 3.

**Algorithm 3**: Pseudocode for the Hybrid FA algorithms (Ezugwu, 2022))

**Input:** Initial schedule $X = (x_1, x_2 \ldots x_n)^t$ at $t = 0$; $x_{cgbest} = \emptyset$; $\gamma = 1$; $\alpha = 0.20$; $\beta_0 = 0.20$; $\lambda = 0.25$; $maxFE$

**Output**: Good quality schedule $x_{cbest}$

1: Define fitness function $f(X)$ based on the objective function $C_{max}$: maximum completion time
2: Generate initial population of fireflies $x_i (i = 1,2, \ldots .n)$
3: Light intensity $I_i$ of firefly $x_i$ is determined using $f(x_i)$
4: while $(t < maxFE)$ do
5:    for $i = 1 : n$ for all $n$ firefly
6:       for $j = 1 : i$ +1 for $n$ all fireflies
7:          $\alpha = AlphaNew()$;   // determine a new value of $\alpha$
8:          Stage 1: Solve $(x_i)$    // find $x_i$ using equation (11)
9:          Stage 2: Solve $(x_j)$    // find $x_j$ using equations (9) and (11)
10:          Find $C_{max}(x_i)$ that are associated with $x_i$ and $x_j$
11:          $EvalFA(x_i, f(x_i))$ // evaluate $x$ based on $f(x_i)$ associated with $x_i$ and $x_j$
12:          If $(f(x_i) < f(x_j))$ Then
13:            Move $x_i$ towards $x_j$ in $d$-dimensions
14:          end if
15:          $X' = Mutate(X)$ //using local search scheme generate a new schedule from $X$
16:          Calculate new fitness values for all fireflies
17:          Update firefly light intensity $I_i$
18:    end for
19:    end for
20: $RankFA(x_t, f(X'))$;   // rank the fireflies according to their function values

21: $x_{cgbest} = FindCurrentgBestFA(x_t, f(\mathbf{X'}))$;   // determine the current global best solution
22: $x_{t+1} = MoveFA(x_t)$;   // Vary attractiveness with distance $r$ via $exp(-\gamma r)$
23: *end while*

### 4.3. Initial solution

In the case of the UPMSP, the initial solution for the metaheuristic algorithm is typically generated randomly. This involves ordering the sequence of jobs and assigning them to machines (Ezugwu, 2018). The initial population size for the candidate swarm or organisms is set to be equal to the number of solutions in the population. It is important to note that the quality of the initial population has a significant impact on the performance of representative algorithms as population-based metaheuristics. A good initial population increases the algorithm's chances of discovering promising areas within the search space and provides diversity to avoid premature convergence (Garey and Johnson, 1979; Ezugwu and Akutsah, 2018; Ezugwu, 2022).

In this study, a maximum of 200 candidate particle swarms or organisms were used as the population size to solve the UPMSP. This population size selection allows for a sufficient number of solutions to be evaluated and explored during the optimization process.

### 4.4. Mutation based local search improvement schemes

To enhance the quality of solutions obtained in parallel machine scheduling, a mutation-based local search procedure is incorporated into the representative algorithm. This procedure aims to improve the generated schedule quality by utilizing three neighborhood structures or mutation operators (Ezugwu, 2019; 2022). The mutation-based local search acts as an improvement phase, focusing on minimizing the total completion time ($C_{max}$) of the current schedule. Three neighborhood operators are utilized in this local search improvement scheme: swap sequence mutation, insertion sequence mutation, and reverse sequence mutation operations. The swap operator involves swapping the schedules of two selected jobs, denoted as $i$ and $j$, from two different machines, $k$ and $l$, to interchange their positions. It should be noted that the selection of jobs and machines for swapping is done randomly. The insertion sequence mutation operation is implemented in two instances: intra-machine insertion and inter-machine insertion. In the case of intra-machine insertion, a random job, $i$, from schedule $X$ is selected and inserted before a randomly chosen job, $j$, within the same schedule (Ezugwu, 2019). By incorporating these mutation-based local search procedures and employing the swap, insertion, and reverse sequence mutation operators, the representative algorithm aims to enhance the quality of the obtained schedules in parallel machine scheduling problems.

In the case of inter-machine insertion, the insertion operation involves removing a job, denoted as $i$, from one machine, $k$, and inserting it into another machine, $l$. This operation facilitates the transfer of a job between machines, potentially improving the overall schedule. The revert operator is another component of the local search improvement procedure. It selects a random job from schedule $X$ and replaces its sub-schedule with the reversed order of the original sub-schedule. This reversal of the job's sub-schedule aims to explore alternative arrangements and potentially enhance the schedule quality.

The local search improvement operation is performed iteratively for a specified number, Γ, of times for each of the three neighborhood operators. This iterative process allows for multiple iterations of improvement within the local search phase. For a detailed depiction of the steps involved in each of the local search improvement processes, refer to Algorithm Listing 3. This algorithm provides a comprehensive illustration of the specific steps and procedures employed during the local search improvement phase.

**Algorithm 3**: Pseudocode for the mutation based local search improvement scheme (Ezugwu, 2019; 2022)
**Input**: solution consisting of schedules $X$ and two machines $l$ and $m$
**Output**: Improved solution or good-quality schedule X′ for the representative optimization methods
1: $P\_swap = 0.2$;
2: $P\_revert = 0.5$;
3: $P\_insert = 1 - (P\_swap + P\_revert)$;
4: $P = [P\_swap, P\_revert, P\_insert]$;
5: $SCHEME = SSPF(P)$; where $SSPF(.)$ denotes scheme selection probability function
6: *for each SCHEME do*
7:     Γ = 1
8:     Randomly select two machines $l$ and $m$ such that $l \neq m$
9:     Randomly select two jobs $i_l$ and $j_m$
10:    *while* Γ ≤ 3 *do*
11:       *if* Γ = 1 and $k \neq l$ *then*
12:          X′= **ApplySwap**($X$, $i_l$, $l$, $i_k$, $k$)
12:          *if* Γ = 1 and $k \neq m$ *then*
12:             X′= **ApplySwap**($X$, $i_m$, $m$, $i_k$, $k$)
12:          *end if*
12:       *end if*
13:       *else if* Γ = 2 and $k \neq l$ *then*
14:          X′= **ApplyInsert**($X$, $i_l$, $l$, $k$)
13:          *if* Γ = 2 and $k \neq m$ *then*
14:             X′= **ApplyInsert**($X$, $i_m$, $m$, $k$)
13:          *end if*
12:       *end if*
15:       *else if* Γ == 3 *then*
16:          X′= **ApplyRevert**($X$, $i_\pi, \pi, i_k, k$)// $\pi$ = randomly selected machine ($l$ or $m$)
17:       *end if*
18:       *if* X′ *is better than* X *then*
19.          X ← X′
20:       *else*
21:       *end if*
22:    Γ = Γ + 1
23:    *end while*
24: *end for*

In Algorithm listing 3, the value of Γ, which represents the number of random moves, is dependent on the available machines. To fine-tune the performance of each representative algorithm, the values of $M$ were chosen from the intervals [0.5m, 0.9m]. For a similar experimental configuration that resulted in good quality solutions, please refer to (Ezugwu and Akutsah, 2018; Ezugwu, 2022).

In the algorithm, the parameter $\pi$ represents the selected machine ($l$ or $m$), distinct from the intervened machine $k$. The parameter $p$ is determined using the proportionate fitness selection, also known as roulette wheel selection. This selection mechanism associates a probability of selection with each solution or swarm particle in the population based on their fitness level. If $\delta_i$ represents the fitness of candidate solution $i$ in the population, its probability of being selected is expressed according to equation (16).

$$p_i = \frac{\delta_i}{\sum_{j=1}^{\psi} \delta_j}, j = 1, 2, \ldots, \psi$$

(16)

In the optimization process, the population size, denoted by ψ, plays a crucial role. A metaheuristic algorithm is deemed advantageous and superior when it achieves a balance between candidate solution selection and exploration of new solution regions within the solution landscape. It is important to note that the selection task in most metaheuristic algorithms has the potential to transform an exploratory search process into a hill climber by rejecting all exploratory search solutions. Achieving this balance is essential for generating high-quality solutions.

## 5. Experiments

In this section, we present the experimental configurations, dataset description, and discussion of the experimental results. Three experiments were conducted, and their results are discussed individually. Experiment 1 focuses on evaluating nine state-of-the-art metaheuristic algorithms: FA, ACO, GA, IWO, ABC, BA, DE, PSO, and TLBO. The results of this experiment are presented, by comparing the performance of these algorithms. In Experiment 2, we examine the influence of population size on the FA algorithm, which demonstrated the best performance in the previous experiment. The study takes into account the variations in population sizes and the number of iterations or function evaluations used in previous literature. To validate the FA algorithm's performance, we specifically investigate its sensitivity to the initialization parameter of population size. A fixed number of 500,000 function evaluations was used to test the FA method in the optimization process. Experiment 3 involves hybridization approaches that combine the FA algorithm with other optimization techniques, including FADE, FAPSO, FAABC, FATLBO, and FAIWO. The results obtained from these hybrid approaches are presented, along with the average CPU time consumed by each algorithm and their respective percentage deviations.

### 5.1 Experimental Configuration

All the algorithms were implemented in MATLAB, utilizing a computational PC equipped with an Intel(R) Core(TM) i7-7700 CPU operating at 3.60GHz and 16 GB of RAM. This configuration provided the necessary computational power and memory resources for executing the algorithms and conducting the experiments effectively.

To evaluate the performance of the algorithms, each problem instance was tested with fifteen different scenarios. The quality of a solution was determined by calculating the average value of the makespan $C_{max}$. The termination condition for the execution of the algorithms was set at a total of 500,000 function evaluations for all the problem instances. This allowed recording the best fitness function value obtained, which in this case is the average $C_{max}$. To ensure fair comparison and consistency, the number of function evaluations was chosen as the termination criterion instead of a specific number of iterations. The parameter settings used for the algorithms were kept the same as in the original algorithm implementations from the source references. Additionally, the number of function evaluations and swarm population size were kept constant for all the algorithms.

The parameter settings used for each algorithm implementation were as follows: swarm population size ($\psi$) = 40, $\gamma = 1$, $\beta_0 = 2$, maximum number of function evaluations (maxFE) = 500,000, $\alpha = 0.2$, and uniform mutation rate (um) = 2. These parameter values were selected based on previous experimental trials and were consistent with similar studies discussed in previous works (Ezugwu & Akutsah, 2018; Ezugwu, 2019; Ezugwu, Ezugwu, Adeleke, & Viriri, 2018; Ezugwu, 2022).

### 5.2 Benchmark instances

The experimental dataset comprises a total of 1620 test instances, involving different combinations of machines and jobs. Specifically, the dataset includes 2, 4, 6, 8, 10, and 12 machines, with corresponding job numbers of 20, 40, 60, 80, 100, and 120. These instances were used to evaluate the performance of the algorithms under investigation. The benchmark instance are available at http://www.schedulingResearch.com.

### 5.3. Evaluation metrics

To evaluate the quality of solutions obtained by the representative algorithms, the average percentage deviation ($\rho$) from the lower bound was used. The average percentage deviation of the makespan $C_{max}$ for each algorithm was recorded. The calculation of the average percentage deviation (APD) is performed using the following equation (17) as presented in(Ezugwu, 2022).

$$\rho = \frac{C_{max}(optimizer) - LB}{LB} \times 100\% \tag{17}$$

The average percentage deviation ($\delta$) of other competing algorithms from the FA algorithm was calculated using FA as the control algorithm. This calculation compares the average $C_{max}$ obtained by the competing algorithms with the average $C_{max}$ obtained by the FA algorithm. The average percentage deviation is calculated as shown in equation (18), as proposed by Ezugwu (2022):

$$\delta = \frac{C_{max}(optimizer) - C_{max}(FA)}{C_{max}(FA)} \times 100\% \tag{18}$$

The lower bound (LB) for each test instance was determined using the methods presented in equations (20) through (21), as follows:

- The first lower bound, denoted by LB1, is calculated as the ratio between the sum of the minimum adjusted processing time's matrix and the number of machines. Mathematically, it can be expressed as shown in equation (19) (Ezugwu and Akutsah, 2018; Ezugwu, 2022).

$$LB1 = \frac{\sum_{j=1}^{n} \min_{\substack{k=1,2,\ldots,m \\ i=1,2,\ldots,n}}[AP_{i,j,k}]}{m}$$

(19)

- The second lower bound, denoted by LB2, is calculated as the maximum value in the adjusted processing times matrix divided by the minimum value in the adjusted processing times matrix for each job. Mathematically, it can be expressed as shown in equation (20) (Ezugwu and Akutsah, 2018; Ezugwu, 2022).

$$LB2 = \max_{j=1,2,..n} \left\{ \min_{\substack{k=1,2,\ldots,m \\ i=1,2,\ldots,n}}[AP_{i,j,k}] \right.$$

(20)

- The final lower bound is determined as the maximum value between the first lower bound (LB1) and the second lower bound (LB2). It is calculated and expressed as shown in equation (21) (Ezugwu and Akutsah, 2018; Ezugwu, 2022).

$$LB = max(LB1, LB2)$$

(21)

### 5.4. Experiment 1: Evaluation of representative metaheuristic algorithms

To assess the performance of the representative optimization algorithms, a comprehensive set of test problem instances was considered. The instances consisted of different combinations of machines and jobs, ranging from 2 to 12 machines and 20 to 120 jobs. Similar implementation approaches were followed as in previous studies (Ezugwu and Akutsah, 2018; Ezugwu, 2019; Ezugwu, 2022). Each algorithm was applied to solve 15 instances for each combination of machines and jobs. Table 1 presents the average $C_{max}$ results and standard deviations obtained by the algorithms. The results clearly demonstrate that the FA algorithm outperformed all other representative optimization algorithms in terms of average makespan ($C_{max}$) and standard deviation (Std) values. The performance of the FA algorithm remained superior across different problem instances, particularly for a large number of jobs (80, 100, and 120) and 8, 10, and 12 machine combinations.

It is worth noting that although the FA algorithm showed superior performance in terms of average $C_{max}$, its Std values were comparatively higher than those of the ABC, BA, DE, and TLBO algorithms. However, the FA algorithm consistently maintained its performance superiority even with a larger combination of jobs and machines, as evident from the numerical computations in Table 1. In a more general sense, considering the primary objective function of achieving minimum $C_{max}$, the FA algorithm demonstrated a high capability in finding the optimal solutions among all tested combinations of machines and jobs. However, the IWO, BA, and TLBO algorithms also yielded better $C_{max}$ results compared to the other algorithms, namely PSO, ACO, GA, ABC, and DE. The ACO algorithm exhibited the lowest performance, obtaining the highest $C_{max}$ values and Std values.

In order to align our experiments with the objective of optimizing the assignment of jobs to machines and reducing the overall makespan of the scheduling problem, which can contribute to the goal of reducing energy consumption and aligning with the sustainable development goals (SDGs), we conducted further investigations to evaluate the performance of the FA algorithm. Considering that the FA algorithm performed well in the previous experiments, we decided to explore its potential in hybridizing with other metaheuristics discussed in this paper.

In this section of the paper, we focused on two main aspects. Firstly, we investigated the population size that would be most suitable for the FA algorithm and its hybrid variants to effectively handle the scheduling problem under consideration. The goal was to determine an optimal population size that balances solution quality and computational efficiency. After identifying the appropriate population size, we proceeded to explore the hybrid implementations of the FA algorithm with other metaheuristics on the test instances of the UPMSP. The objective was to assess the effectiveness of combining the strengths of different algorithms in improving the overall performance and solution quality.

By conducting these critical experiments, we aimed to provide insights into the potential enhancements and synergies that can be achieved through hybridization. The findings from these experiments can contribute to the development of more powerful and efficient optimization approaches for solving the scheduling problem at hand. In the subsequent sections, we will present the experimental setup, discuss the results obtained, and analyze the performance of the hybrid approaches in comparison to the standalone FA algorithm. This comprehensive evaluation will shed light on the capabilities and potential of hybrid metaheuristics in tackling the UPMSP and contribute to advancing the state-of-the-art in scheduling optimization.

Table 1: Average $C_{max}$ and $Std$ values obtained by each algorithm for all tested problem instances (Ezugwu, 2022)

| m | n | LB | ACO | | GA | | IWO | | FA | | ABC | | BA | | DE | | PSO | | TLBO | |
|---|---|---|---|---|---|---|---|---|---|---|---|---|---|---|---|---|---|---|---|---|
| | | Avg $C_{max}$ | Avg $C_{max}$ | Std | Avg $C_{max}$ | Std | Avg $C_{max}$ | Std | Avg $C_{max}$ | Std | Avg $C_{max}$ | Std | Avg $C_{max}$ | Std | Avg $C_{max}$ | Std | Avg $C_{max}$ | Std | Avg $C_{max}$ | Std |
| 2 | 20 | 1185.83 | 1346.87 | 27.59 | 1229.93 | 47.44 | 1208.2 | 33.5 | 1189.60 | 34.71 | 1190 | 35.34 | 1194.93 | 37.78 | 1193 | 31.39 | 1217.27 | 34.34 | 1190.67 | 35.92 |
| | 40 | 2344.7 | 2833.8 | 36.57 | 2433.53 | 42.14 | 2412.27 | 41.47 | 2384.13 | 27.81 | 2443.07 | 38.55 | 2414.27 | 42.09 | 2403.73 | 34.48 | 2468.8 | 70.33 | 2400.13 | 33.47 |
| | 60 | 3510.17 | 4322.87 | 52.97 | 3702.53 | 61.76 | 3629.73 | 47.86 | 3588.40 | 45.40 | 3706.93 | 37.2 | 3638.53 | 47.81 | 3640.67 | 40.94 | 3739.07 | 114.92 | 3633.67 | 40.57 |
| | 80 | 4664.83 | 5823 | 58.17 | 4910.67 | 66.29 | 4825.67 | 69.62 | 4778.20 | 59.42 | 4943.93 | 55.03 | 4840.33 | 65.12 | 4864.73 | 58.77 | 4896.6 | 100.07 | 4859.2 | 55.95 |
| | 100 | 5819.23 | 7304.53 | 93.97 | 6134.4 | 74.44 | 6032.2 | 59.06 | 5955.33 | 66.76 | 6201.8 | 59.42 | 6044.6 | 56.82 | 6112.93 | 57.67 | 6233.67 | 160.18 | 6094.87 | 53.33 |
| | 120 | 7008.03 | 8826.87 | 78.95 | 7374.73 | 81.78 | 7265.47 | 73.9 | 7175.00 | 63.18 | 7473.47 | 60.02 | 7271.53 | 82.71 | 7387.07 | 69.23 | 7378.4 | 120.48 | 7361.07 | 66.69 |
| 4 | 20 | 560.83 | 717.33 | 33.61 | 591.53 | 18.6 | 571.27 | 19.25 | 559.33 | 14.06 | 565.07 | 10.13 | 570.6 | 14.07 | 566.4 | 13.37 | 562.8 | 11.41 | 565.53 | 12.74 |
| | 40 | 1101.88 | 1538.87 | 57.15 | 1202.47 | 37.39 | 1175.8 | 31.82 | 1140.67 | 20.71 | 1186.73 | 14.15 | 1171.4 | 17.31 | 1164.13 | 13.82 | 1207.4 | 41.86 | 1174.2 | 19.61 |
| | 60 | 1650.73 | 2401.73 | 98.69 | 2022.53 | 774.55 | 1793 | 39.52 | 1720.00 | 23.84 | 1814.67 | 16.58 | 1783.6 | 32.04 | 1779.67 | 19.19 | 1844.47 | 53.87 | 1790.13 | 18.52 |
| | 80 | 2201.48 | 3259.4 | 125.65 | 2417.73 | 34.15 | 2371.27 | 27.62 | 2309.87 | 24.95 | 2453.93 | 20.41 | 2406.07 | 20.16 | 2404.8 | 13.8 | 2476.8 | 111 | 2402.87 | 26.4 |
| | 100 | 2740.7 | 4111.4 | 174.35 | 3090.13 | 118.38 | 2942.4 | 59.14 | 2871.20 | 34.52 | 3073.4 | 23.47 | 2976.53 | 49.96 | 3032.33 | 28.07 | 3121.73 | 133.41 | 3013.6 | 21.07 |
| | 120 | 3291.2 | 4956.73 | 216.11 | 3692.4 | 89.73 | 3547.27 | 61.83 | 3442.87 | 34.45 | 3702.07 | 24.09 | 3593.73 | 36.75 | 3661.13 | 29.45 | 3700.67 | 160.73 | 3651.07 | 33.18 |
| 6 | 20 | 362.4 | 523.47 | 24.61 | 424.6 | 22.59 | 407.07 | 17.37 | 393.67 | 9.23 | 392 | 4.64 | 399.67 | 10.97 | 396.13 | 6.83 | 404.8 | 13.33 | 398.93 | 6.1 |
| | 40 | 716.56 | 1136.87 | 77.75 | 806.6 | 53.5 | 798.93 | 19.45 | 761.87 | 15.98 | 795.13 | 12.12 | 783.87 | 15.56 | 774.33 | 8.38 | 782.93 | 42.56 | 773.07 | 10.48 |
| | 60 | 1071.48 | 1771.33 | 178.3 | 1205.8 | 36.09 | 1201.07 | 25.26 | 1142.60 | 25.31 | 1211.2 | 11.47 | 1202.13 | 25.8 | 1196.53 | 18.41 | 1210.8 | 56.67 | 1189.93 | 18.88 |
| | 80 | 1429.12 | 2442.93 | 110.44 | 1633.2 | 23.14 | 1607.67 | 31.58 | 1530.53 | 19.48 | 1641.4 | 15.2 | 1604.47 | 22.95 | 1617.87 | 14.08 | 1651.67 | 81.61 | 1607.33 | 19.21 |
| | 100 | 1783.03 | 3051 | 198.22 | 2059 | 49.78 | 2003.8 | 40.08 | 1922.47 | 42.67 | 2063.8 | 12.64 | 2011.07 | 24.04 | 2044.93 | 23.96 | 2073.8 | 85.81 | 2026.6 | 14.72 |
| | 120 | 2137.6 | 3797.07 | 222.51 | 2499.87 | 71.23 | 2380.13 | 41.14 | 2301.33 | 36.34 | 2492 | 15.31 | 2431.27 | 35.24 | 2474.47 | 29.22 | 2513.87 | 116.15 | 2463.6 | 24.34 |
| 8 | 20 | 267.23 | 440.47 | 34.51 | 302.4 | 9.44 | 296.87 | 22.58 | 291.93 | 7.20 | 291.53 | 4.67 | 294.4 | 10.83 | 294.87 | 8.94 | 290.93 | 5.61 | 293.4 | 7.02 |
| | 40 | 529.76 | 940.73 | 77.55 | 606.53 | 30.35 | 579.13 | 34.78 | 548.80 | 9.14 | 586.73 | 6.13 | 573.93 | 16.97 | 576.27 | 7.88 | 598.2 | 33.88 | 576.47 | 14.77 |
| | 60 | 791.74 | 1492.27 | 93.45 | 915.2 | 25.75 | 902.8 | 19 | 873.00 | 20.01 | 917.4 | 10.02 | 904.53 | 19.29 | 901.93 | 9.93 | 920.53 | 35.44 | 903.2 | 14.88 |
| | 80 | 1053.08 | 1952.67 | 149.28 | 1213.73 | 57.7 | 1208.53 | 47.45 | 1159.87 | 28.08 | 1227.8 | 10.33 | 1211.73 | 26.17 | 1221.73 | 15.54 | 1230.07 | 52.01 | 1220.13 | 14.33 |
| | 100 | 1315.38 | 2520.13 | 209.12 | 1533.07 | 67.82 | 1521.53 | 54.47 | 1441.40 | 33.48 | 1549.6 | 11.27 | 1531.8 | 26.32 | 1550.07 | 21.79 | 1580.33 | 71.34 | 1542.4 | 18.08 |
|    | 120 | 1580.23 | 3075.07 | 185.27 | 1865.4  | 98.84 | 1834.73 | 83.71 | 1746.73 | 37.93 | 1875.07 | 11.67 | 1853.8  | 32.24 | 1897.67 | 19.91 | 1911.47 | 98.75 | 1880.87 | 26.51 |
|    | 20  | 210.85  | 375.4   | 17.27  | 223.27  | 37.12 | 214.13  | 10.88 | 193.87  | 5.10  | 197.33  | 2.47  | 197.53  | 4.98  | 195.73  | 3.95  | 184.2   | 3.9   | 194.73  | 3.53  |
|    | 40  | 419.89  | 800.07  | 55.2   | 489.07  | 46.67 | 468.33  | 31.25 | 435.53  | 10.72 | 461.73  | 4.46  | 463.93  | 9.38  | 453.47  | 7.91  | 463.93  | 43.42 | 457.2   | 9.65  |
| 10 | 60  | 625.56  | 1244.8  | 94.52  | 721.67  | 30.11 | 716.4   | 26.03 | 690.13  | 26.78 | 722.33  | 5.86  | 715.87  | 21.23 | 726.4   | 19.15 | 722.13  | 45.3  | 713.07  | 18.35 |
|    | 80  | 835.12  | 1762.4  | 125.08 | 1013.87 | 41.47 | 973.8   | 48.09 | 938.60  | 20.18 | 988.53  | 6.5   | 992.73  | 23.23 | 987.13  | 15.15 | 1007.4  | 60.35 | 990.6   | 13.75 |
|    | 100 | 1041.54 | 2256.67 | 135.97 | 1267.47 | 34.08 | 1222.87 | 28.21 | 1174.53 | 26.40 | 1246.87 | 8.89  | 1251.4  | 27.91 | 1255.27 | 8.54  | 1243.6  | 60.8  | 1252.33 | 17.01 |
|    | 120 | 1249.07 | 2700.67 | 174.19 | 1494.33 | 47.55 | 1479    | 55.76 | 1425.40 | 43.90 | 1503.4  | 13.54 | 1505.2  | 30.78 | 1528.07 | 20.33 | 1558.4  | 77.11 | 1522.47 | 18.26 |
|    | 20  | 174.59  | 343.4   | 24.46  | 194.47  | 10.93 | 189.2   | 10.37 | 182.87  | 7.37  | 185.47  | 2.85  | 187.8   | 5.47  | 182.87  | 4.31  | 173.2   | 3.93  | 184.73  | 4.45  |
|    | 40  | 346.93  | 717.13  | 48.46  | 433.2   | 35.13 | 429.33  | 26.22 | 400.00  | 7.81  | 409.93  | 4.93  | 420.73  | 12.53 | 401.67  | 6.65  | 395.8   | 6.18  | 407.47  | 9.34  |
| 12 | 60  | 519.21  | 1117    | 54.44  | 635.8   | 41.39 | 602.6   | 69.12 | 577.87  | 27.28 | 598     | 3.98  | 608.33  | 23.74 | 608.27  | 10.89 | 611.47  | 24.54 | 597.33  | 13.11 |
|    | 80  | 690.47  | 1528.87 | 93.62  | 840.87  | 69.43 | 834.87  | 35.12 | 795.53  | 22.79 | 828.4   | 7.44  | 832.87  | 17.67 | 835.4   | 13.99 | 847.13  | 54.26 | 828.07  | 14.61 |
|    | 100 | 863.53  | 1938.13 | 138.82 | 1063.73 | 44.73 | 1036    | 49.07 | 1002.27 | 29.99 | 1048.8  | 5.87  | 1054.93 | 28.08 | 1073.33 | 19.86 | 1060.07 | 53.94 | 1055.93 | 27.43 |
|    | 120 | 1034.79 | 2388    | 87.1   | 1288.93 | 45.19 | 1251.87 | 23.13 | 1212.07 | 43.41 | 1261.53 | 8.42  | 1269.4  | 38.4  | 1294.07 | 20.11 | 1272.27 | 52.86 | 1287.87 | 23.06 |

### 5.3.1. Experiment 2: Evaluation of influence of population size on FA

The influence of population size on the FA optimization algorithm was evaluated to understand its impact on the algorithm's performance. In this evaluation, different population sizes were tested to observe their effects on the convergence behavior and the quality of solutions obtained. The population size is an important parameter in population-based metaheuristic algorithms like FA (Ezugwu, Adeleke, Akinyelu, & Viriri, 2020; Li, Liu, & Yang, 2020; Ezugwu, 2022). It determines the diversity and exploration-exploitation trade-off within the population. A larger population size generally allows for more exploration of the search space, increasing the chances of finding better solutions. However, it may also lead to slower convergence and higher computational costs.

To evaluate the influence of population size, a range of population sizes was considered, such as 20, 30, 40, and 50. The FA algorithm was applied to solve a set of benchmark problem instances using each population size, and the results were compared and analyzed. The evaluation focused on two main aspects: the convergence behavior and the quality of solutions. The convergence behavior was assessed by monitoring the progress of the algorithm's objective function ($C_{max}$) over number of function evaluations. The rate of convergence, stability, and the number of function evaluation needed to reach a certain level of convergence were considered.

The quality of solutions was evaluated by comparing the average $C_{max}$ values obtained by the algorithm with different population sizes. Additionally, statistical analysis techniques, such as hypothesis testing or confidence intervals, could be employed to determine if the differences in performance were statistically significant. However, this type of analysis was beyond the scope of the current paper. Based on the evaluation, it was observed that the population size had a notable influence on the FA algorithm's performance. Larger population sizes generally led to better exploration of the search space, resulting in improved solutions. However, the convergence rate was slower with larger populations, as more individuals needed to be evaluated and updated in each iteration.

On the other hand, smaller population sizes showed faster convergence but had a higher risk of getting trapped in local optima. They may also exhibit reduced exploration capabilities, limiting their ability to discover globally optimal solutions. Therefore, selecting an appropriate population size for the FA algorithm is a trade-off between exploration and exploitation. It depends on the complexity of the problem, the computational resources available, and the desired balance between solution quality and convergence speed.

In summary, the evaluation of the influence of population size on the FA optimization algorithm highlighted the importance of selecting an appropriate population size that balances exploration and exploitation. The results provided insights into the trade-offs involved and can guide practitioners in determining the optimal population size for their specific problem instances. More so, the summary of the average results for the influence of varying population sizes on different number of machines and jobs scales is presented in Table 2.

Table 2: Influence of population size on the FA algorithm.

| m | n | Metrics | FA/20 | FA/30 | FA/40 |
|---|---|---|---|---|---|
| 2 | 20 | Mean | 1188.87 | 1186 | 1179.80 |
| | | Best | 1143 | 1136 | 1133 |
| | | Std | 35.29 | 33.59 | 33.67 |
| | | Median | 1188 | 1186 | 1187 |
| | | Worst | 1254 | 1247 | 1238 |
| | 40 | Mean | 2386.40 | 2387.07 | 2316.60 |
| | | Best | 2327 | 2325 | 2293 |
| | | Std | 35.64 | 40.42 | 36.86 |
| | | Median | 2382 | 2392 | 2372 |
| | | Worst | 2480 | 2481 | 2452 |
| | 60 | Mean | 3597.67 | 3580.47 | 3561.67 |
| | | Best | 3531 | 3503 | 3520 |
| | | Std | 39.06 | 39.71 | 40.03 |
| | | Median | 3606 | 3586 | 3592 |
| | | Worst | 3655 | 3633 | 3643 |
| | 80 | Mean | 4783.53 | 4777.93 | 4741.60 |
| | | Best | 4686 | 4689 | 4621 |
| | | Std | 54.15 | 56.85 | 52.63 |
| | | Median | 4770 | 4768 | 4734 |
| | | Worst | 4900 | 4902 | 4925 |
| | 100 | Mean | 5965.67 | 5971.33 | 5934.13 |
| | | Best | 5844 | 5856 | 5850 |
| | | Std | 63.01 | 65.99 | 59.05 |
| | | Median | 5967 | 5981 | 5939 |
| | | Worst | 6081 | 6093 | 6053 |
| | 120 | Mean | 7207.60 | 7207.67 | 7123.93 |
| | | Best | 7126 | 7117 | 7018 |
| | | Std | 64.38 | 61.41 | 61.30 |
| | | Median | 7207 | 7215 | 7115 |
| | | Worst | 7320 | 7318 | 7235 |
| 4 | 20 | Mean | 559.80 | 558.80 | 552.33 |
| | | Best | 544 | 537 | 534 |
| | | Std | 11.18 | 11.94 | 9.79 |
| | | Median | 560 | 556 | 550 |
| | | Worst | 583 | 584 | 573 |
| | 40 | Mean | 1150.47 | 1136.07 | 1125.40 |
| | | Best | 1097 | 1102 | 1103 |
| | | Std | 30.86 | 29.88 | 16.28 |
| | | Median | 1147 | 1134 | 1125 |
| | | Worst | 1214 | 1213 | 1163 |
| | 60 | Mean | 1738.87 | 1718.87 | 1720.00 |
| | | Best | 1681 | 1683 | 1683 |
| | | Std | 39.99 | 26.79 | 19.02 |
| | | Median | 1721 | 1709 | 1715 |
| | | Worst | 1814 | 1775 | 1769 |
| | 80 | Mean | 2319.47 | 2314.33 | 2313.00 |
| | | Best | 2269 | 2278 | 2286 |
| | | Std | 27.87 | 19.39 | 13.91 |
| | | Median | 2327 | 2311 | 2312 |
| | | Worst | 2364 | 2350 | 2396 |
| | 100 | Mean | 2891.73 | 2881.53 | 2843.93 |
| | | Best | 2844 | 2849 | 2805 |
| | | Std | 31.41 | 20.68 | 22.51 |
| | | Median | 2896 | 2879 | 2896 |
| | | Worst | 2949 | 2907 | 2948 |
| | 120 | Mean | 3470.33 | 3464.53 | 3413.33 |
| | | Best | 3436 | 3430 | 3329 |
| | | Std | 27.44 | 22.74 | 19.76 |
| | | Median | 3469 | 3461 | 3416 |
| | | Worst | 3519 | 3508 | 3530 |
| | 20 | Mean | 400.80 | 399.40 | 391.13 |
| | | Best | 379 | 388 | 373 |
| | | Std | 15.75 | 7.51 | 6.65 |
| | | Median | 395 | 400 | 390 |
| | | Worst | 446 | 414 | 404 |
| | 40 | Mean | 765.80 | 756.67 | 752 |
| | | Best | 747 | 733 | 727 |
| | | Std | 18.27 | 19.26 | 10.88 |
| | | Median | 764 | 753 | 753 |
| | | Worst | 804 | 800 | 771 |
| | 60 | Mean | 1158.07 | 1146.67 | 1135.13 |
| | | Best | 1120 | 1110 | 1113 |
| | | Std | 22.40 | 25.70 | 12.72 |
| | | Median | 1163 | 1142 | 1132 |
| | | Worst | 1187 | 1201 | 1158 |
| 6 | 80 | Mean | 1553.13 | 1522.73 | 1530.13 |
| | | Best | 1513 | 1493 | 1509 |
| | | Std | 29.97 | 16.53 | 16.38 |
| | | Median | 1550 | 1519 | 1529 |
| | | Worst | 1629 | 1561 | 1564 |
| | 100 | Mean | 1942.60 | 1920.40 | 1919.80 |
| | | Best | 1897 | 1871 | 1888 |
| | | Std | 37.60 | 30.65 | 12.73 |
| | | Median | 1930 | 1918 | 1912 |
| | | Worst | 2028 | 1989 | 2004 |
| | 120 | Mean | 2312.27 | 2310.13 | 2304.60 |
| | | Best | 2269 | 2270 | 2268 |
| | | Std | 33.38 | 37.36 | 17.91 |
| | | Median | 2312 | 2307 | 2302 |
| | | Worst | 2412 | 2410 | 2335 |
| 8 | 20 | Mean | 292.20 | 290.73 | 284.80 |

| | | Best | 282 | 276 | 273 |
| | | Std | 5.51 | 6.60 | 6.68 |
| | | Median | 293 | 293 | 285 |
| | | Worst | 299 | 299 | 298 |
| | 40 | Mean | 553.73 | 548.20 | 541.00 |
| | | Best | 535 | 531 | 529 |
| | | Std | 21.29 | 11.49 | 8.09 |
| | | Median | 546 | 546 | 540 |
| | | Worst | 619 | 572 | 554 |
| | 60 | Mean | 870.53 | 866.93 | 863.40 |
| | | Best | 844 | 847 | 845 |
| | | Std | 15.40 | 17.70 | 10.80 |
| | | Median | 873 | 862 | 862 |
| | | Worst | 894 | 910 | 885 |
| | 80 | Mean | 1177.27 | 1156.20 | 1154.20 |
| | | Best | 1102 | 1110 | 1110 |
| | | Std | 42.43 | 27.19 | 12.31 |
| | | Median | 1179 | 1164 | 1155 |
| | | Worst | 1256 | 1205 | 1179 |
| | 100 | Mean | 1461.73 | 1443.13 | 1435.40 |
| | | Best | 1399 | 1412 | 1402 |
| | | Std | 37.31 | 20.84 | 14.00 |
| | | Median | 1465 | 1440 | 1437 |
| | | Worst | 1525 | 1488 | 1464 |
| | 120 | Mean | 1758.20 | 1743.53 | 1738.47 |
| | | Best | 1708 | 1699 | 1707 |
| | | Std | 41.52 | 24.16 | 14.26 |
| | | Median | 1744 | 1737 | 1739 |
| | | Worst | 1835 | 1789 | 1784 |
| 10 | 20 | Mean | 198.87 | 191.47 | 188.00 |
| | | Best | 189 | 178 | 180 |
| | | Std | 8.21 | 5.99 | 4.68 |
| | | Median | 196 | 191 | 188 |
| | | Worst | 212 | 204 | 196 |
| | 40 | Mean | 435.20 | 434.60 | 425.07 |
| | | Best | 422 | 422 | 413 |
| | | Std | 14.69 | 8.84 | 6.40 |
| | | Median | 431 | 434 | 425 |
| | | Worst | 485 | 446 | 436 |
| | 60 | Mean | 691.53 | 693.60 | 671.80 |
| | | Best | 659 | 648 | 648 |
| | | Std | 21.44 | 34.24 | 9.89 |
| | | Median | 697 | 687 | 665 |
| | | Worst | 728 | 758 | 694 |
| | 80 | Mean | 961.80 | 933.13 | 922.20 |
| | | Best | 898 | 890 | 881 |
| | | Std | 36.30 | 22.47 | 12.67 |
|  |  | Median | 961 | 932 | 920 |
|  |  | Worst | 1037 | 978 | 947 |
|  | 100 | Mean | 1192.40 | 1175.80 | 1171.87 |
|  |  | Best | 1155 | 1147 | 1128 |
|  |  | Std | 28.30 | 19.41 | 7.36 |
|  |  | Median | 1187 | 1176 | 1179 |
|  |  | Worst | 1247 | 1215 | 1193 |
|  | 120 | Mean | 1452.27 | 1413.93 | 1397.00 |
|  |  | Best | 1394 | 1380 | 1354 |
|  |  | Std | 35.45 | 32.01 | 11.45 |
|  |  | Median | 1455 | 1397 | 1411 |
|  |  | Worst | 1537 | 1487 | 1430 |
| 12 | 20 | Mean | 183.47 | 182.67 | 178.67 |
|  |  | Best | 172 | 172 | 172 |
|  |  | Std | 8.58 | 8.03 | 4.59 |
|  |  | Median | 184 | 182 | 178 |
|  |  | Worst | 197 | 197 | 189 |
|  | 40 | Mean | 413.53 | 405.20 | 393.33 |
|  |  | Best | 388 | 389 | 383 |
|  |  | Std | 19.18 | 15.47 | 6.94 |
|  |  | Median | 410 | 401 | 393 |
|  |  | Worst | 454 | 440 | 408 |
|  | 60 | Mean | 582.93 | 573.20 | 564.80 |
|  |  | Best | 546 | 534 | 534 |
|  |  | Std | 32.45 | 26.91 | 5.66 |
|  |  | Median | 592 | 567 | 551 |
|  |  | Worst | 652 | 617 | 609 |
|  | 80 | Mean | 805.20 | 772.00 | 782.67 |
|  |  | Best | 759 | 748 | 728 |
|  |  | Std | 33.87 | 14.60 | 8.29 |
|  |  | Median | 796 | 773 | 775 |
|  |  | Worst | 886 | 802 | 827 |
|  | 100 | Mean | 1016.60 | 999.53 | 997.73 |
|  |  | Best | 966 | 969 | 957 |
|  |  | Std | 25.86 | 21.33 | 8.25 |
|  |  | Median | 1019 | 1001 | 997 |
|  |  | Worst | 1065 | 1037 | 1064 |
|  | 120 | Mean | 1229.47 | 1204.47 | 1184.93 |
|  |  | Best | 1166 | 1168 | 1146 |
|  |  | Std | 34.36 | 23.89 | 14.22 |
|  |  | Median | 1221 | 1207 | 1122 |
|  |  | Worst | 1301 | 1245 | 1273 |

Figure 4 illustrates the average CPU times consumed by the FA algorithm using different population sizes of 20, 30, and 40 for the complete set of test problems. The purpose of analyzing the computational times is to understand the efficiency and runtime performance of the FA algorithm with varying population sizes. Based on the results, it can be observed that the FA algorithm with a population size of 40 (FA40) is the fastest among the three population sizes tested. It achieved the shortest average execution time compared to FA with population sizes of 20 (FA20) and 30 (FA30). The FA algorithm with a population size of 30 came second with the second least execution time.

However, it is worth noting that FA40 recorded the highest CPU time on smaller problem instances, specifically when there were 20 jobs for 4, 6, 8, and 10 machines. This suggests that the larger population size requires more computational resources and may result in longer execution times for smaller problem instances.

The findings from Figure 4 provide insights into the trade-off between population size and computational efficiency. While a larger population size may lead to better solution quality and faster convergence on larger problem instances, it may also incur higher computational costs. On the other hand, a smaller population size may be more efficient for smaller problem instances but might sacrifice solution quality. These results can guide practitioners and researchers in selecting an appropriate population size for the FA algorithm based on the problem size and available computational resources. Moreover, the findings clearly highlights the importance of considering the balance between solution quality and computational efficiency when configuring the population size for metaheuristic algorithms.

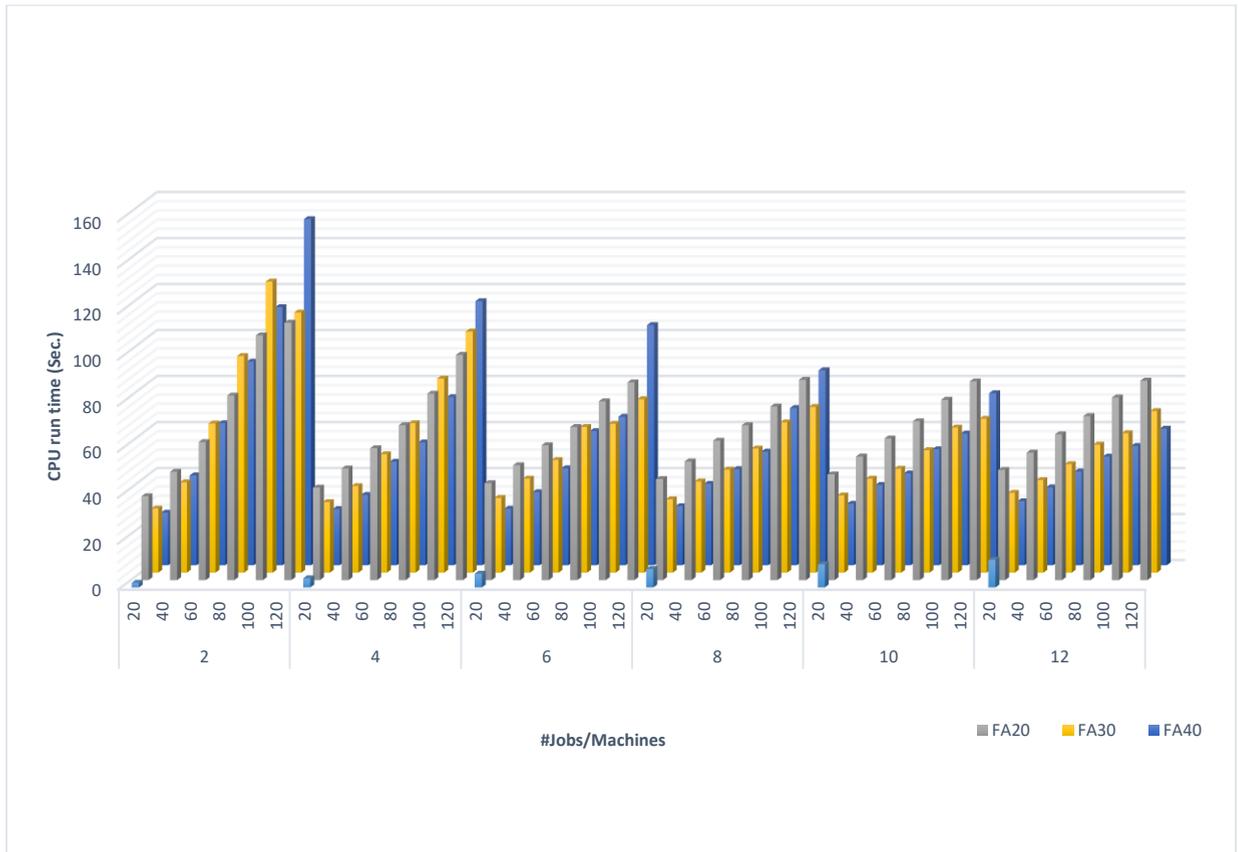

Fig. 4: Average computational time for FA with 20, 30, and 40 population sizes on all test instances

### 5.3.3. Experiment 3: Implementation and evaluation of hybrid FA for UPMSP

For the current set of experiments, we retained the same experimental configuration as in the previous experiments, with some adjustments to the population size and number of function evaluations. Specifically, we increased the population size to 200 and reduced the maximum number of function evaluations to 5000. These adjustments were necessary due to the significant increase in population size compared to the previous experiment, where we used a population size of 40. By reducing the number of function evaluations, we aimed to ensure a reasonable computational time while still allowing sufficient exploration and exploitation of the search space.

Apart from the changes mentioned above, all other parameter configurations remained the same as previously presented in Section 5.1. This consistency in parameter settings ensured a fair comparison between the hybridized FA algorithm and the standalone FA algorithm, as well as consistency with the previous experiments. By maintaining the same experimental setup, we could effectively evaluate the impact of hybridization on the performance and solution quality of the FA algorithm for the UPMSP.

In the following sections, we will describe the results of these experiments and analyze the performance of the hybridized FA algorithm with the adjusted population size. The objective is to assess whether the hybrid approaches can further enhance the performance of the FA algorithm in terms of solution quality, computational efficiency, and its ability to optimize the assignment of jobs to machines in the scheduling problem.

Table 3 presents the best solutions obtained by each of the proposed metaheuristic algorithms. The table is structured with the first column representing the number of machines, followed by the second column indicating the number of jobs (n), and subsequent columns representing the results of each algorithm. Each algorithm is executed on all instances with the given number of machines and jobs, and the objective is to identify the algorithm that achieves the best optimal results in terms of minimizing the makespan.

In table 3, the minimum value represents the best solution found, while the maximum value represents the worst solution. The average of the makespan for $n$ jobs on $m$ machines is also provided, along with the standard deviation. It is important to note that the default parameter configuration of each algorithm is retained as described in the original literature, ensuring consistency and fairness in the comparative analysis.

Analyzing the results in Table 4, we observe that the FADE algorithm performs worse compared to the standard FA algorithm, indicating that the hybridization did not significantly improve its performance. Following FADE, the FAPSO algorithm also shows relatively weaker results compared to the other hybrid algorithms. On the other hand, FAIWO, FAABC, and FATLBO demonstrate highly competitive results compared to the standard FA algorithm. Among these hybrid algorithms, FAIWO emerges as the best-performing algorithm for all the test instances.

It is worth mentioning that as the number of jobs increases, the performance of the standard FA algorithm seems to decline. However, the hybrid algorithms, particularly FAIWO, maintain their competitiveness even with a larger number of jobs. Among the hybrid algorithms, FAPSO appears to be the least performing, indicating that the hybridization of FA with the PSO algorithm may not have been as effective in improving its performance compared to the other hybrid approaches.

These findings demonstrate the impact of hybridization on the performance of the FA algorithm, with FAIWO standing out as the most promising hybrid variant. The results suggest that incorporating the IWO algorithm into the FA algorithm yields improved solutions for the scheduling problem, while other hybrid approaches may not be as effective.

Table 3: Present the Best, Worst, Avg., and Std values obtained by each algorithm for all tested problem instances

| | | FA | | | | FADE | | | | FAPSO | | | | FAABC | | | | FATLBO | | | | FAIWO | | | |
|---|---|---|---|---|---|---|---|---|---|---|---|---|---|---|---|---|---|---|---|---|---|---|---|---|---|
| m | n | Best | Worst | Avg. | Std | Best | Worst | Avg. | Std | Best | Worst | Avg. | Std | Best | Worst | Avg. | Std | Best | Worst | Avg. | Std | Best | Worst | Avg. | Std |
| 2 | 20 | 1133 | 1257 | 1189.6000 | 34.7065 | 1158 | 1256 | 1204.8670 | 32.4475 | 1280 | 1398 | 1336.4670 | 36.8857 | 1133 | 1238 | 1177.9333 | 34.5780 | 1136 | 1235 | 1178.4667 | 32.7062 | 1136 | 1238 | 1177.5333 | 34.2863 |
| | 40 | 2354 | 2451 | 2384.1333 | 27.8128 | 2406 | 2553 | 2487.2670 | 37.4766 | 2721 | 2875 | 2803.8670 | 46.0153 | 2297 | 2441 | 2364.8000 | 32.7462 | 2291 | 2438 | 2368.8000 | 36.8514 | 2290 | 2443 | 2368.0000 | 35.8050 |
| | 60 | 3501 | 3652 | 3588.4000 | 45.3995 | 3732 | 3850 | 3790.9330 | 39.3364 | 4231 | 4353 | 4293.1330 | 39.4731 | 3531 | 3649 | 3594.0000 | 38.7243 | 3533 | 3647 | 3598.4000 | 35.9121 | 3505 | 3619 | 3570.2667 | 34.0325 |
| | 80 | 4687 | 4900 | 4778.2000 | 59.4177 | 5004 | 5201 | 5090.6000 | 54.4634 | 5683 | 5837 | 5753.3330 | 44.9725 | 4770 | 4951 | 4839.4667 | 53.1935 | 4738 | 4971 | 4831.2667 | 60.4335 | 4711 | 4902 | 4771.9333 | 53.9344 |
| | 100 | 5848 | 6116 | 5955.3333 | 66.7615 | 6255 | 6462 | 6389.9330 | 56.8977 | 7087 | 7288 | 7213.5330 | 56.1870 | 5995 | 6191 | 6091.8667 | 53.9667 | 6000 | 6161 | 6086.4667 | 52.7986 | 5897 | 6109 | 5995.1333 | 59.1727 |
| | 120 | 7089 | 7299 | 7175.0000 | 63.1800 | 7645 | 7840 | 7720.4000 | 55.6440 | 8620 | 8853 | 8724.6670 | 73.4396 | 7287 | 7519 | 7393.7333 | 65.8239 | 7297 | 7515 | 7389.6000 | 68.7030 | 7163 | 7370 | 7253.0667 | 67.3927 |
| 4 | 20 | 535 | 587 | 559.3333 | 14.0645 | 553 | 581 | 569.5333 | 7.8273 | 653 | 742 | 709.0667 | 23.1932 | 535 | 568 | 548.3333 | 10.5469 | 535 | 569 | 549.4000 | 8.63382 | 535 | 574 | 549.0000 | 10.2539 |
| | 40 | 1106 | 1179 | 1140.6667 | 20.7146 | 1181 | 1226 | 1201.333 | 14.7487 | 1479 | 1569 | 1519.867 | 30.6475 | 1106 | 1150 | 1128.6000 | 12.8497 | 1104 | 1146 | 1124.8667 | 11.9872 | 1104 | 1144 | 1120.5333 | 12.4031 |
| | 60 | 1685 | 1774 | 1720.0000 | 23.8357 | 1815 | 1880 | 1849.0000 | 15.8520 | 2156 | 2416 | 2302.333 | 70.1220 | 1705 | 1764 | 1732.7333 | 15.0782 | 1699 | 1756 | 1727.5333 | 16.0038 | 1681 | 1737 | 1705.3333 | 17.3603 |
| | 80 | 2258 | 2370 | 2309.8667 | 24.9510 | 2479 | 2516 | 2498.067 | 11.5107 | 2984 | 3309 | 3139.800 | 104.6028 | 2336 | 2395 | 2363.4667 | 16.9532 | 2340 | 2399 | 2361.800 | 15.4003 | 2283 | 2327 | 2304.1333 | 13.8763 |
| | 100 | 2810 | 2939 | 2871.2000 | 34.5216 | 3098 | 3182 | 3135.800 | 23.1893 | 3766 | 4170 | 3946.600 | 99.7832 | 2962 | 3025 | 2990.5333 | 20.1135 | 2947 | 3011 | 2990.0667 | 18.8052 | 2316 | 2938 | 2830.8000 | 204.0725 |
| | 120 | 3372 | 3520 | 3442.8667 | 34.4549 | 3756 | 3831 | 3782.067 | 20.9369 | 4492 | 5027 | 4704.333 | 142.2240 | 3607 | 3685 | 3635.0000 | 21.4842 | 3603 | 3691 | 3633.3333 | 20.7490 | 3498 | 3580 | 3524.0000 | 19.3723 |
| 6 | 20 | 380 | 413 | 393.6667 | 9.2324 | 378 | 406 | 393.2667 | 7.7870 | 491 | 558 | 524.8000 | 18.1746 | 377 | 395 | 387.1333 | 5.7305 | 377 | 395 | 387.6667 | 6.54290 | 378 | 398 | 387.2000 | 5.4798 |
| | 40 | 743 | 797 | 761.8667 | 15.9815 | 784 | 815 | 794.6667 | 8.1475 | 992 | 1226 | 1123.867 | 51.1899 | 732 | 765 | 745.6000 | 8.7489 | 737 | 768 | 747.8000 | 9.41276 | 733 | 764 | 744.8667 | 8.5679 |
| | 60 | 1111 | 1199 | 1142.6000 | 25.3065 | 1198 | 1240 | 1221.267 | 11.5045 | 1598 | 1853 | 1740.667 | 70.6265 | 1123 | 1154 | 1140.0667 | 9.2386 | 1130 | 1156 | 1140.6667 | 7.57502 | 1098 | 1129 | 1116.1333 | 8.2624 |
| | 80 | 1508 | 1577 | 1530.5333 | 19.4821 | 1642 | 1674 | 1654.267 | 8.5479 | 2151 | 2522 | 2320.600 | 118.0495 | 1552 | 1585 | 1566.8000 | 10.4758 | 1547 | 1582 | 1567.6000 | 10.29424 | 1504 | 1550 | 1527.2667 | 1527.2667 |
| | 100 | 1859 | 1985 | 1922.4667 | 42.6663 | 2072 | 2112 | 2090.000 | 10.3026 | 2623 | 3183 | 2945.333 | 174.0938 | 1977 | 2020 | 1999.8000 | 12.6107 | 1982 | 2007 | 1993.1333 | 8.53452 | 1915 | 1942 | 1929.9333 | 9.5951 |
| | 120 | 2243 | 2374 | 2301.3333 | 36.3449 | 2483 | 2552 | 2517.067 | 19.7170 | 3497 | 3887 | 3638.867 | 93.8250 | 2410 | 2456 | 2431.5333 | 14.2223 | 2394 | 2451 | 2427.7333 | 13.70332 | 2310 | 2376 | 2340.7333 | 17.5722 |
| 8 | 20 | 273 | 305 | 291.9333 | 7.1959 | 278 | 299 | 288.2000 | 5.3211 | 363 | 490 | 431.4667 | 44.1521 | 272 | 293 | 281.8000 | 6.6138 | 274 | 297 | 284.9333 | 5.82441 | 273 | 295 | 281.3333 | 6.1140 |
| | 40 | 535 | 566 | 548.8000 | 9.1355 | 575 | 596 | 584.0667 | 5.0067 | 877 | 1052 | 942.2000 | 43.7904 | 532 | 547 | 541.1333 | 4.4700 | 530 | 560 | 540.1333 | 7.78154 | 523 | 552 | 535.6667 | 7.8710 |
| | 60 | 842 | 917 | 873.0000 | 20.0071 | 903 | 922 | 913.9333 | 6.8187 | 1362 | 1555 | 1473.3330 | 56.2236 | 851 | 882 | 866.5333 | 9.2572 | 848 | 884 | 863.8000 | 8.84954 | 537 | 864 | 831.4667 | 81.9258 |
| | 80 | 1108 | 1205 | 1159.8667 | 28.0837 | 1226 | 1246 | 1234.400 | 6.5444 | 1828 | 2137 | 1980.400 | 90.2535 | 1160 | 1183 | 1171.5333 | 7.2788 | 1161 | 1187 | 1172.5333 | 7.84553 | 843 | 1167 | 1117.2000 | 77.2900 |
| | 100 | 1397 | 1509 | 1441.4000 | 33.4809 | 1542 | 1580 | 1563.400 | 10.2316 | 2170 | 2706 | 2440.867 | 151.5123 | 1478 | 1519 | 1457.9333 | 10.0655 | 1485 | 1521 | 1501.2667 | 10.8789 | 1148 | 1466 | 1427.4000 | 78.0255 |
| | 120 | 1697 | 1823 | 1746.7333 | 37.9256 | 1873 | 1911 | 1889.133 | 11.1090 | 2707 | 3196 | 2956.933 | 132.3596 | 1804 | 1849 | 1833.4000 | 12.2171 | 1817 | 1848 | 1833.000 | 9.41883 | 1467 | 1785 | 1745.2667 | 77.6811 |
| 10 | 20 | 186 | 202 | 193.8667 | 5.0972 | 186 | 203 | 195.3333 | 3.8483 | 343 | 398 | 364.8667 | 15.0564 | 175 | 195 | 185.2667 | 5.1195 | 180 | 195 | 185.5333 | 3.62268 | 175 | 193 | 185.0667 | 4.5898 |
| | 40 | 422 | 454 | 435.5333 | 10.7229 | 454 | 471 | 461.4667 | 4.5492 | 774 | 891 | 827.1333 | 31.9975 | 419 | 435 | 426.6000 | 4.3720 | 419 | 435 | 426.1333 | 4.37199 | 415 | 432 | 423.4667 | 4.7339 |
| | 60 | 641 | 728 | 690.1333 | 26.7792 | 716 | 730 | 722.6000 | 4.1542 | 1107 | 1374 | 1225.7330 | 69.9045 | 661 | 684 | 673.1333 | 7.2296 | 665 | 680 | 674.6000 | 4.54816 | 644 | 664 | 654.4667 | 5.7180 |

| | | | | | | | | | | | | | | | | | | | | | | | |
|---|---|---|---|---|---|---|---|---|---|---|---|---|---|---|---|---|---|---|---|---|---|---|---|
| | 80 | 898 | 972 | 938.6000 | 20.1771 | 971 | 990 | 984.4667 | 5.2490 | 1528 | 1844 | 1712.0670 | 90.6134 | 919 | 953 | 937.9333 | 10.3955 | 922 | 947 | 936.5333 | 8.22771 | 885 | 944 | 905.0667 | 13.0847 |
| | 100 | 1136 | 1225 | 1174.5333 | 26.4031 | 1234 | 1259 | 1246.4000 | 7.3465 | 1897 | 2287 | 2133.4670 | 109.2683 | 1191 | 1223 | 1204.8000 | 8.2652 | 1188 | 1211 | 1200.6000 | 7.28795 | 1136 | 1177 | 1153.5333 | 11.8072 |
| | 120 | 1370 | 1514 | 1425.4000 | 43.9005 | 1495 | 1520 | 1507.9330 | 8.6476 | 2277 | 2784 | 2571.0000 | 127.2714 | 1445 | 1488 | 1466.5333 | 11.1795 | 1452 | 1490 | 1470.1333 | 9.54588 | 1397 | 1432 | 1417.9333 | 9.6988 |
| 12 | 20 | 172 | 199 | 182.8667 | 7.3666 | 178 | 190 | 184.9333 | 3.3481 | 247 | 374 | 334.9333 | 40.7264 | 165 | 180 | 173.1333 | 4.5649 | 167 | 182 | 174.9333 | 4.30061 | 170 | 181 | 175.2667 | 3.0347 |
| | 40 | 390 | 416 | 400.0000 | 7.8102 | 392 | 411 | 399.6000 | 5.1796 | 639 | 775 | 710.7333 | 43.6880 | 378 | 398 | 388.9333 | 4.7729 | 375 | 396 | 386.0000 | 5.74456 | 381 | 396 | 388.8000 | 4.1952 |
| | 60 | 538 | 647 | 577.8667 | 27.2813 | 585 | 603 | 597.2000 | 4.3785 | 1016 | 1175 | 1112.133 | 48.8275 | 547 | 571 | 558.0000 | 5.8064 | 546 | 572 | 557.9333 | 6.80826 | 536 | 550 | 543.8667 | 4.3403 |
| | 80 | 753 | 841 | 795.5333 | 22.7937 | 815 | 838 | 824.7333 | 6.1466 | 1320 | 1595 | 1466.0670 | 75.6170 | 778 | 801 | 788.4667 | 6.8229 | 777 | 797 | 787.6000 | 5.24813 | 753 | 772 | 764.4000 | 6.6740 |
| | 100 | 953 | 1075 | 1002.2667 | 29.9868 | 1031 | 1054 | 1041.7330 | 6.2389 | 1685 | 1992 | 1861.5330 | 97.5294 | 1002 | 1027 | 1013.2000 | 7.5517 | 1001 | 1028 | 1012.4000 | 9.21799 | 967 | 987 | 976.4000 | 5.6290 |
| | 120 | 1144 | 1335 | 1212.0667 | 43.4091 | 1247 | 1277 | 1258.2000 | 7.6644 | 2059 | 2516 | 2279.9330 | 109.2059 | 1221 | 1247 | 1230.7333 | 6.8917 | 1218 | 1240 | 1228.8000 | 6.43872 | 1161 | 1199 | 1182.9333 | 11.3356 |
| Average | | 1629.917 | 1731.17 | 1673.8185 | 27.9553 | 1759.444 | 1813.25 | 1785.7800 | 15.2256 | 2241.75 | 2505.75 | 2376.8310 | 74.2084 | 1678.694 | 1734.056 | 1703.6241 | 15.8322 | 1678.194 | 1732.444 | 1703.9296 | 15.58358 | 1596.333 | 1699.889 | 1664.5944 | 70.9015 |

Tables 4 show the results of the average percentage deviation $\rho$ of all algorithms from the lower bound, while Table 3 provides the average percentage deviation $\delta$ of other competing algorithms from the FA. In this instance, the FA and the computed lower bound were used as the control algorithms. The obtained results were calculated using equations 21 and 22. The analysis of the results presented in Table 4 reveals that FAPSO has exhibited the worst deviation results in terms of both $\rho$ and $\delta$ values. On the other hand, FAIWO has demonstrated its superiority by outperforming other algorithms in terms of both $\rho$ and $\delta$ values. Following FAIWO, FAABC and FATLBO have also shown competitive performance, ranking second and third, respectively.

It is worth noting that FAIWO has outperformed the standard FA algorithm in terms of the computed $\rho$ results, although it fell behind in comparison to the computed lower bound solution. This indicates that FAIWO has achieved better results compared to FA, even though it may not have reached the theoretical lower bound.

From this experiment, we can deduce that only FAABC, FATLBO, and FAIWO can be considered the best-performing hybrid methods when compared to FA. These algorithms have produced superior results and demonstrated effective performance. On the other hand, FAPSO and FADE have been identified as the worst and least efficient among the other hybrid optimization methods considered in this study.

Table 4: Average for δ and ρ values of representative algorithms for all test instances

| m | n | FADE Best | δ | ρ | FAPSO Best | δ | ρ | FAABC Best | δ | ρ | FATLBO Best | δ | ρ | FAIWO Best | δ | ρ |
|---|---|---|---|---|---|---|---|---|---|---|---|---|---|---|---|---|
| 2 | 20 | 1204.8667 | -0.0128 | -0.0161 | 1336.4667 | -0.1235 | -0.1271 | 1177.9333 | 0.0098 | 0.0067 | 1178.4667 | 0.0094 | 0.0062 | 1177.5300 | 0.0101 | 0.0070 |
|   | 40 | 2487.2667 | -0.0433 | -0.0608 | 2803.8667 | -0.1761 | -0.1959 | 2364.8000 | 0.0081 | -0.0086 | 2368.8000 | 0.0064 | -0.0103 | 2368.0000 | 0.0068 | -0.0100 |
|   | 60 | 3790.9333 | -0.0564 | -0.0800 | 4293.1333 | -0.1964 | -0.2231 | 3594.0000 | -0.0016 | -0.0239 | 3598.4000 | -0.0028 | -0.0251 | 3570.2667 | 0.0051 | -0.0171 |
|   | 80 | 5090.6000 | -0.0654 | -0.0913 | 5753.3333 | -0.2041 | -0.2334 | 4839.4667 | -0.0128 | -0.0375 | 4831.2667 | -0.0111 | -0.0357 | 4771.9333 | 0.0013 | -0.0230 |
|   | 100 | 6389.9333 | -0.0730 | -0.0981 | 7213.5333 | -0.2113 | -0.2397 | 6091.8667 | -0.0229 | -0.0469 | 6086.4667 | -0.0220 | -0.0459 | 5995.1333 | -0.0067 | -0.0302 |
|   | 120 | 7720.4000 | -0.0760 | -0.1017 | 8724.6667 | -0.2160 | -0.2450 | 7393.7333 | -0.0305 | -0.0550 | 7389.6000 | -0.0299 | -0.0545 | 7253.0667 | -0.0109 | -0.0350 |
| 4 | 20 | 569.5333 | -0.0182 | -0.0156 | 709.0667 | -0.2677 | -0.2646 | 548.3333 | 0.0197 | 0.0222 | 549.4000 | 0.0178 | 0.0203 | 549.0000 | 0.0185 | 0.0210 |
|   | 40 | 1201.3333 | -0.0532 | -0.0903 | 1519.8667 | -0.3324 | -0.3796 | 1128.6000 | 0.0106 | -0.0243 | 1124.8667 | 0.0139 | -0.0209 | 1120.5333 | 0.0177 | -0.0169 |
|   | 60 | 1849.0000 | -0.0750 | -0.1201 | 2302.3333 | -0.3386 | -0.3948 | 1732.7333 | -0.0074 | -0.0497 | 1727.5333 | -0.0044 | -0.0465 | 1705.3333 | 0.0085 | -0.0331 |
|   | 80 | 2498.0667 | -0.0815 | -0.1347 | 3139.8000 | -0.3593 | -0.4262 | 2363.4667 | -0.0232 | -0.0736 | 2361.8000 | -0.0225 | -0.0728 | 2304.1333 | 0.0025 | -0.0467 |
|   | 100 | 3135.8000 | -0.0922 | -0.1442 | 3946.6000 | -0.3745 | -0.4400 | 2990.5333 | -0.0416 | -0.0912 | 2990.0667 | -0.0414 | -0.0910 | 2830.8000 | 0.0141 | -0.0329 |
|   | 120 | 3782.0667 | -0.0985 | -0.1492 | 4704.3333 | -0.3664 | -0.4294 | 3635.0000 | -0.0558 | -0.1045 | 3633.3333 | -0.0553 | -0.1040 | 3524.0000 | -0.0236 | -0.0707 |
| 6 | 20 | 393.2667 | 0.0010 | -0.0852 | 524.8000 | -0.3331 | -0.4482 | 387.1333 | 0.0166 | -0.0684 | 387.6667 | 0.0152 | -0.0699 | 387.2000 | 0.0164 | -0.0687 |
|   | 40 | 794.6667 | -0.0431 | -0.1090 | 1123.8667 | -0.4751 | -0.5683 | 745.6000 | 0.0214 | -0.0405 | 747.8000 | 0.0185 | -0.0436 | 744.8667 | 0.0223 | -0.0395 |
|   | 60 | 1221.2667 | -0.0688 | -0.1398 | 1740.6667 | -0.5234 | -0.6245 | 1140.0667 | 0.0022 | -0.0640 | 1140.6667 | 0.0017 | -0.0646 | 1116.1333 | 0.0232 | -0.0417 |
|   | 80 | 1654.2667 | -0.0808 | -0.1575 | 2320.6000 | -0.5162 | -0.6238 | 1566.8000 | -0.0237 | -0.0963 | 1567.6000 | -0.0242 | -0.0969 | 1527.2667 | 0.0021 | -0.0687 |
|   | 100 | 2090.0000 | -0.0871 | -0.1722 | 2945.3333 | -0.5321 | -0.6519 | 1999.8000 | -0.0402 | -0.1216 | 1993.1333 | -0.0368 | -0.1178 | 1929.9333 | -0.0039 | -0.0824 |
|   | 120 | 2517.0667 | -0.0937 | -0.1775 | 3638.8667 | -0.5812 | -0.7024 | 2431.5333 | -0.0566 | -0.1375 | 2427.7333 | -0.0549 | -0.1357 | 2340.7333 | -0.0171 | -0.0950 |
| 8 | 20 | 288.2000 | 0.0128 | -0.0785 | 431.4667 | -0.4780 | -0.6162 | 281.8000 | 0.0347 | -0.0545 | 284.9333 | 0.0240 | -0.0663 | 281.3333 | 0.0363 | -0.0528 |
| | 40 | 584.0667 | -0.0643 | -0.1025 | 942.2000 | -0.7168 | -0.7786 | 541.1333 | 0.0140 | -0.0215 | 540.1333 | 0.0158 | -0.0196 | 535.6667 | 0.0239 | -0.0112 |
| | 60 | 913.9333 | -0.0469 | -0.1543 | 1473.3333 | -0.6877 | -0.8609 | 866.5333 | 0.0074 | -0.0945 | 863.8000 | 0.0105 | -0.0910 | 831.4667 | 0.0476 | -0.0503 |
| | 80 | 1234.4000 | -0.0643 | -0.1722 | 1980.4000 | -0.7074 | -0.8805 | 1171.5333 | -0.0101 | -0.1125 | 1172.5333 | -0.0109 | -0.1134 | 1117.2000 | 0.0368 | -0.0608 |
| | 100 | 1563.4000 | -0.0846 | -0.1886 | 2440.8667 | -0.6934 | -0.8557 | 1457.9333 | -0.0115 | -0.1375 | 1501.2667 | -0.0415 | -0.1413 | 1427.4000 | 0.0097 | -0.0851 |
| | 120 | 1889.1333 | -0.0815 | -0.1955 | 2956.9333 | -0.6928 | -0.8711 | 1833.4000 | -0.0496 | -0.1602 | 1833.0000 | -0.0494 | -0.1600 | 1745.2667 | 0.0008 | -0.1045 |
| 10 | 20 | 195.3333 | -0.0076 | 0.0736 | 364.8667 | -0.8820 | -0.7307 | 185.2667 | 0.0444 | 0.1213 | 185.5333 | 0.0430 | 0.1201 | 185.0667 | 0.0454 | 0.1223 |
| | 40 | 461.4667 | -0.0595 | -0.0990 | 827.1333 | -0.8991 | -0.9697 | 426.6000 | 0.0205 | -0.0160 | 426.1333 | 0.0216 | -0.0149 | 423.4667 | 0.0277 | -0.0085 |
| | 60 | 722.6000 | -0.0470 | -0.1551 | 1225.7333 | -0.7761 | -0.9594 | 673.1333 | 0.0246 | -0.0760 | 674.6000 | 0.0225 | -0.0784 | 654.4667 | 0.0517 | -0.0462 |
| | 80 | 984.4667 | -0.0489 | -0.1788 | 1712.0667 | -0.8241 | -1.0501 | 937.9333 | 0.0007 | -0.1231 | 936.5333 | 0.0022 | -0.1215 | 905.0667 | 0.0357 | -0.0838 |
| | 100 | 1246.4000 | -0.0612 | -0.1967 | 2133.4667 | -0.8164 | -1.0483 | 1204.8000 | -0.0258 | -0.1568 | 1200.6000 | -0.0222 | -0.1527 | 1153.5333 | 0.0179 | -0.1075 |
| | 120 | 1507.9333 | -0.0579 | -0.2072 | 2571.0000 | -0.8037 | -1.0582 | 1466.5333 | -0.0289 | -0.1741 | 1470.1333 | -0.0314 | -0.1770 | 1417.9333 | 0.0052 | -0.1352 |
| 12 | 20 | 184.9333 | -0.0113 | -0.0593 | 334.9333 | -0.8316 | -0.9188 | 173.1333 | 0.0532 | 0.0084 | 174.9333 | 0.0434 | -0.0019 | 175.2667 | 0.0416 | -0.0039 |
| | 40 | 399.6000 | 0.0010 | -0.1518 | 710.7333 | -0.7768 | -1.0487 | 388.9333 | 0.0277 | -0.1211 | 386.0000 | 0.0350 | -0.1126 | 388.8000 | 0.0280 | -0.1207 |
| | 60 | 597.2000 | -0.0335 | -0.1502 | 1112.1333 | -0.9246 | -1.1419 | 558.0000 | 0.0344 | -0.0747 | 557.9333 | 0.0345 | -0.0746 | 543.8667 | 0.0588 | -0.0475 |
| | 80 | 824.7333 | -0.0367 | -0.1945 | 1466.0667 | -0.8429 | -1.1234 | 788.4667 | 0.0089 | -0.1419 | 787.6000 | 0.0100 | -0.1407 | 764.4000 | 0.0391 | -0.1071 |
| | 100 | 1041.7333 | -0.0394 | -0.2064 | 1861.5333 | -0.8573 | -1.1557 | 1013.2000 | -0.0109 | -0.1733 | 1012.4000 | -0.0101 | -0.1724 | 976.4000 | 0.0258 | -0.1307 |
| | 120 | 1258.2000 | -0.0381 | -0.2159 | 2279.9333 | -0.8810 | -1.2032 | 1230.7333 | -0.0154 | -0.1894 | 1228.8000 | -0.0138 | -0.1875 | 1182.9333 | 0.0240 | -0.1432 |
| | Avg. | 64288.0667 | -1.8869 | -4.5764 | 85565.9333 | -20.2191 | -24.4888 | 61330.4667 | -0.1096 | -2.7117 | 61341.4667 | -0.1394 | -2.7146 | 59925.3967 | 0.6425 | -1.8601 |

Note: the negative sign means smallest solution (makespan) among all compared algorithms.

With the algorithmic convergence analysis, we employed the option of evaluating each of the algorithm based on the number of iterations consumed instead of the previously utilized number of function evaluations. Ultimately, the choice between iteration and function evaluation depends on the specific problem, the nature of the data, and the desired outcomes. Often, a combination of both approaches can be used to leverage their respective advantages and create more effective and efficient solutions. Therefore, we had based our convergence analysis on some of the advantages of using iteration such as:

- Flexibility: Iteration allows for greater flexibility in solving problems compared to relying solely on function evaluations. It enables you to incorporate complex decision-making, conditionals, and dynamic behavior within the loop structure. Furthermore, this approach aligns perfectly with the issue outlined in the paper.
- Step-by-step processing: Iteration allows you to process data or perform operations incrementally, step by step. This can be useful when dealing with large datasets or performing computationally intensive tasks, as you can process small portions of the data at a time, conserving memory and reducing the overall computational load.
- Problem-specific optimizations: In certain cases, you may be able to optimize the iterative solution to take advantage of specific characteristics of the problem. By tailoring the iteration to exploit patterns or properties of the data, you can often achieve better performance or more efficient solutions compared to a generic function evaluation approach.

To provide further evidence of the superior performance of the individual hybrid algorithms, convergence performance graphs were generated for all the tested algorithms. These algorithms, which are highly competitive optimization methods, were examined separately throughout the experimentation process. The graphs, depicted in Figures 5 to 12, showcase the convergence of each algorithm based on their best $C_{max}$ (smallest) values. The experiments were conducted with varying numbers of machines and jobs, and a fixed number of 500 iterations. The results clearly demonstrate that the FAIWO, FAABC, and FATLBO algorithms exhibit the lowest makespan and most favorable convergence curves among all the algorithms tested under the same experimental settings and conditions. This confirms their superiority over the other algorithms. Furthermore, upon analyzing the convergence curves in more detail, several additional observations can be made, such as:

- The convergence curves of the FAIWO, FAABC, and FATLBO hybrid algorithms vividly depict the impact of the mutation-based local search improvement scheme. This scheme plays a significant role in augmenting the diversity features of the algorithms during the solution search phase. As a result, it contributes to the algorithms' ability to converge towards optimal or near-optimal solutions more effectively. The influence of this scheme is evident in the convergence patterns exhibited by these hybrid algorithms.
- Based on the convergence curves obtained, it is evident that the FAIWO algorithm surpasses other competitive hybrid approaches, including the FAABC, FAPSO, BA, and FATLBO algorithms, when it comes to solving large instances of the test problem. The superior performance of the FAIWO algorithm is clearly demonstrated by its convergence behavior, indicating its effectiveness and efficiency in finding optimal or near-optimal solutions for such challenging instances.

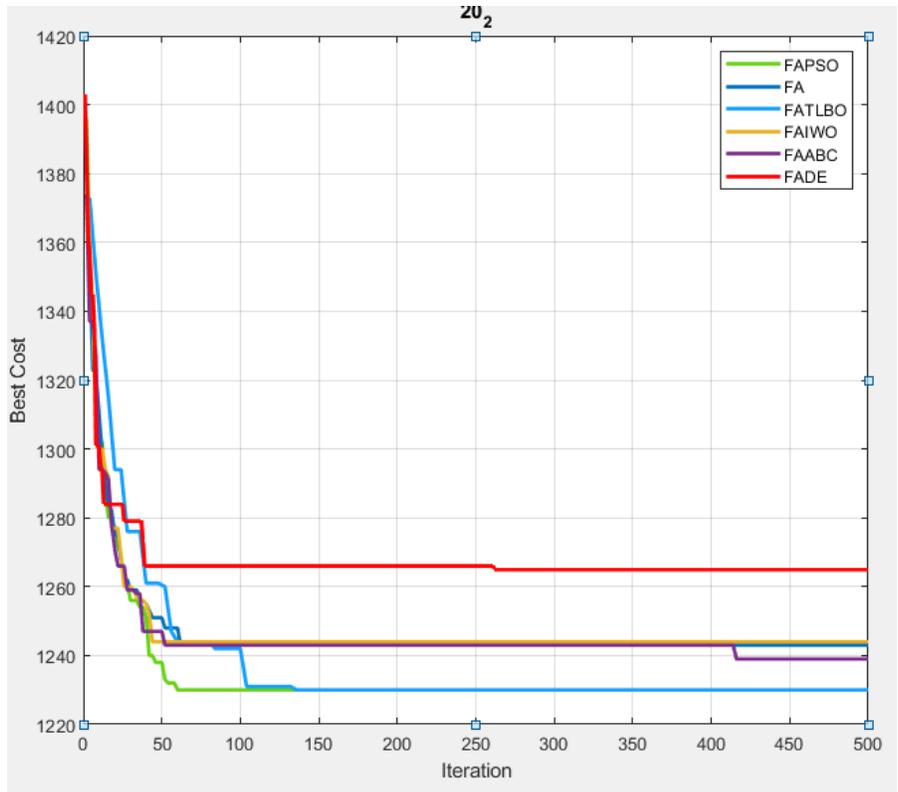

Fig. 5. The convergence graph of the test algorithms on 2 machines and 20 jobs

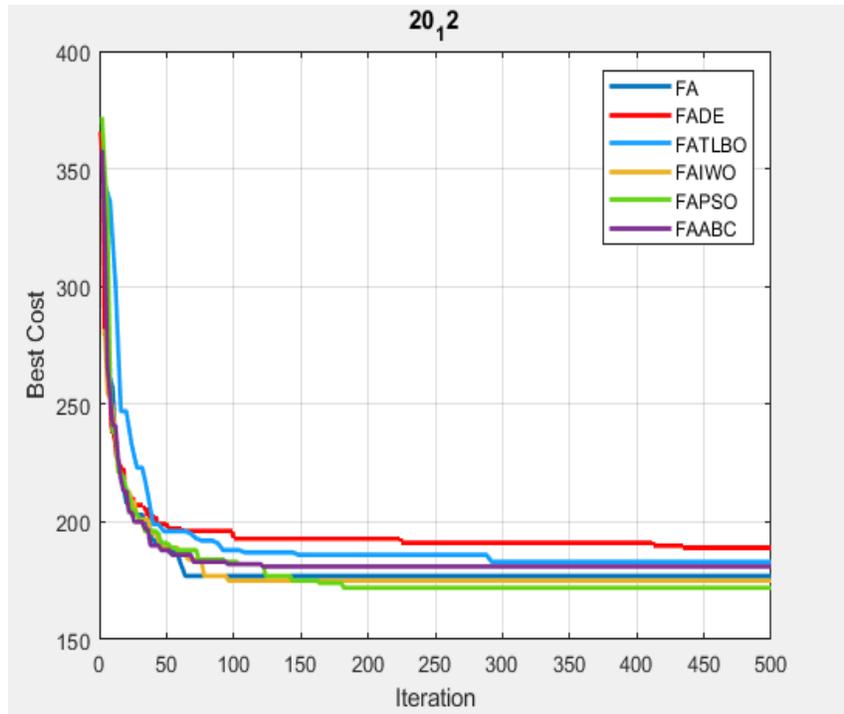

Fig. 6. The convergence graph of the test algorithms on 12 machines and 20 jobs

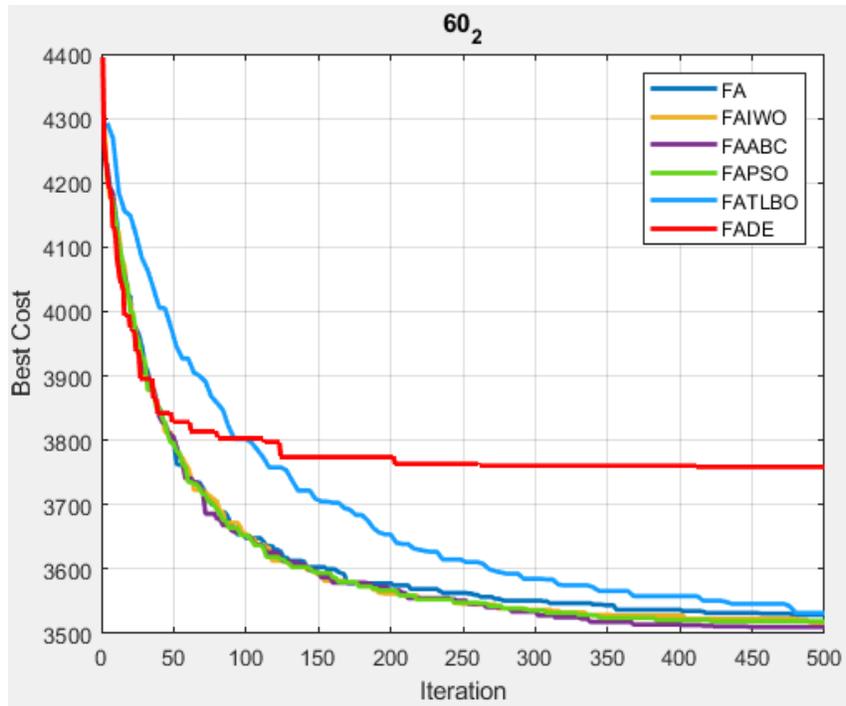

Fig. 7. The convergence graph of the test algorithms on 2 machines and 60 jobs

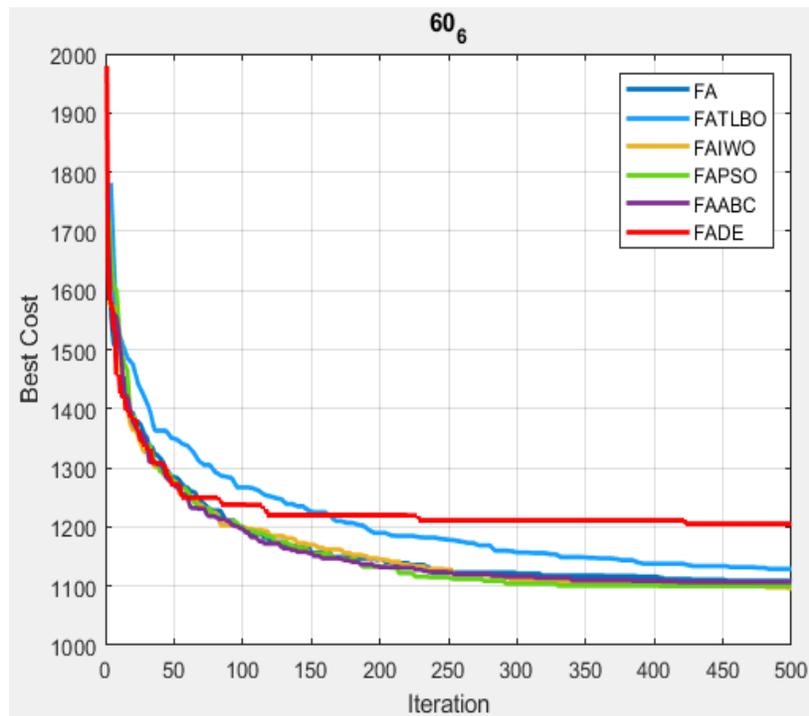

Fig. 8. The convergence graph of the test algorithms on 6 machines and 60 jobs

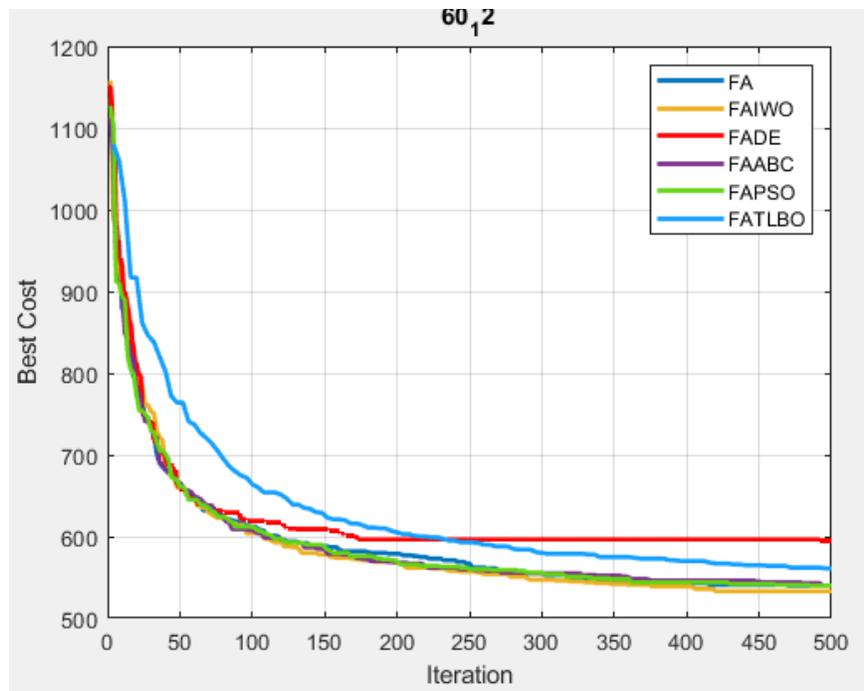

Fig. 9. The convergence graph of the test algorithms on 12 machines and 60 jobs

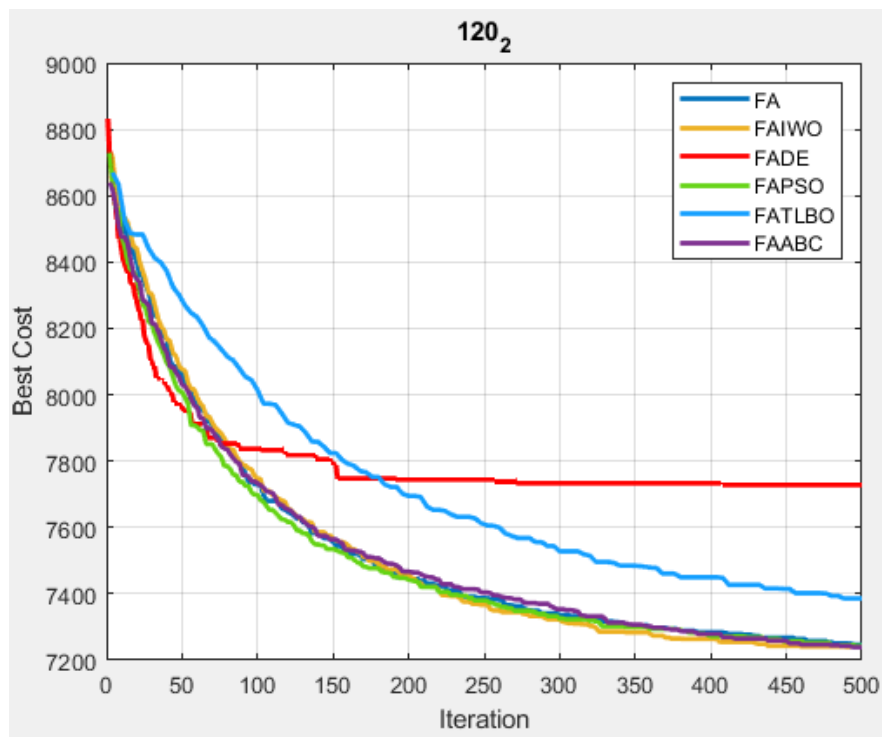

Fig. 10. The convergence graph of the test algorithms on 2 machines and 120 jobs

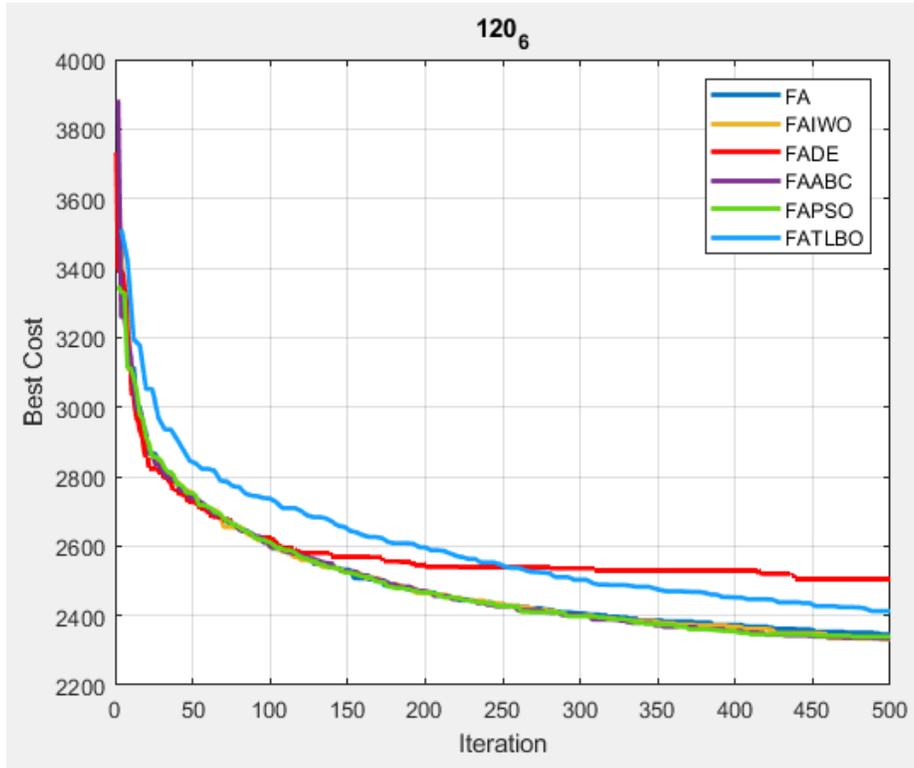

Fig. 11. The convergence graph of the test algorithms on 6 machines and 120 jobs

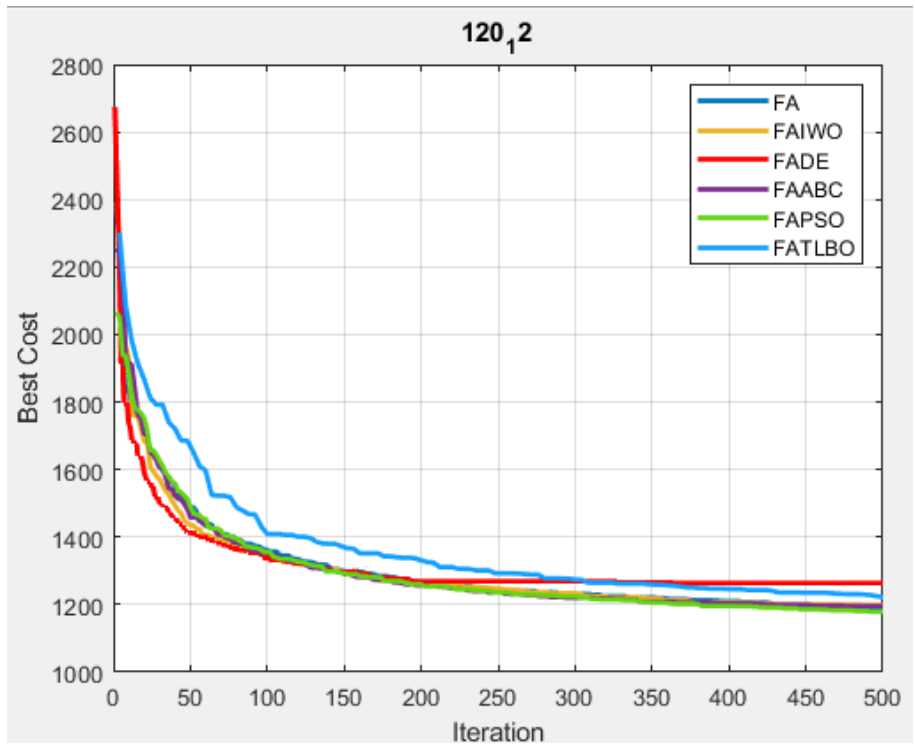

Fig. 12. The convergence graph of the test algorithms on 12 machines and 120 jobs

Furthermore, based on the experiment results carried out to analyze the average execution times consumed by each hybrid methods, the FAPSO stands out as the most efficient algorithm, displaying a quick convergence towards its optimal point within the search space (See Figure 13). Following closely is the FADE algorithm, which also exhibits promising performance results. However, both FAPSO and FADE fall short in terms of achieving the best solution regarding the targeted average minimum $C_{max}$ values. Consequently, our focus shifts to analyzing three remaining algorithms: FAABC, FATLBO, and FAIWO. Among them, FA shows the weakest performance. Moreover, part of the primary objective of the paper is to also propose a novel method that enhances the convergence speed of the FA algorithm.

Upon analysis, FAABC, FATLBO, and FAIWO consistently demonstrate more promising results in both achieving the best solution and exhibiting efficiency within the search space. These algorithms show potential for improving the overall performance and effectiveness of the FA optimization process.

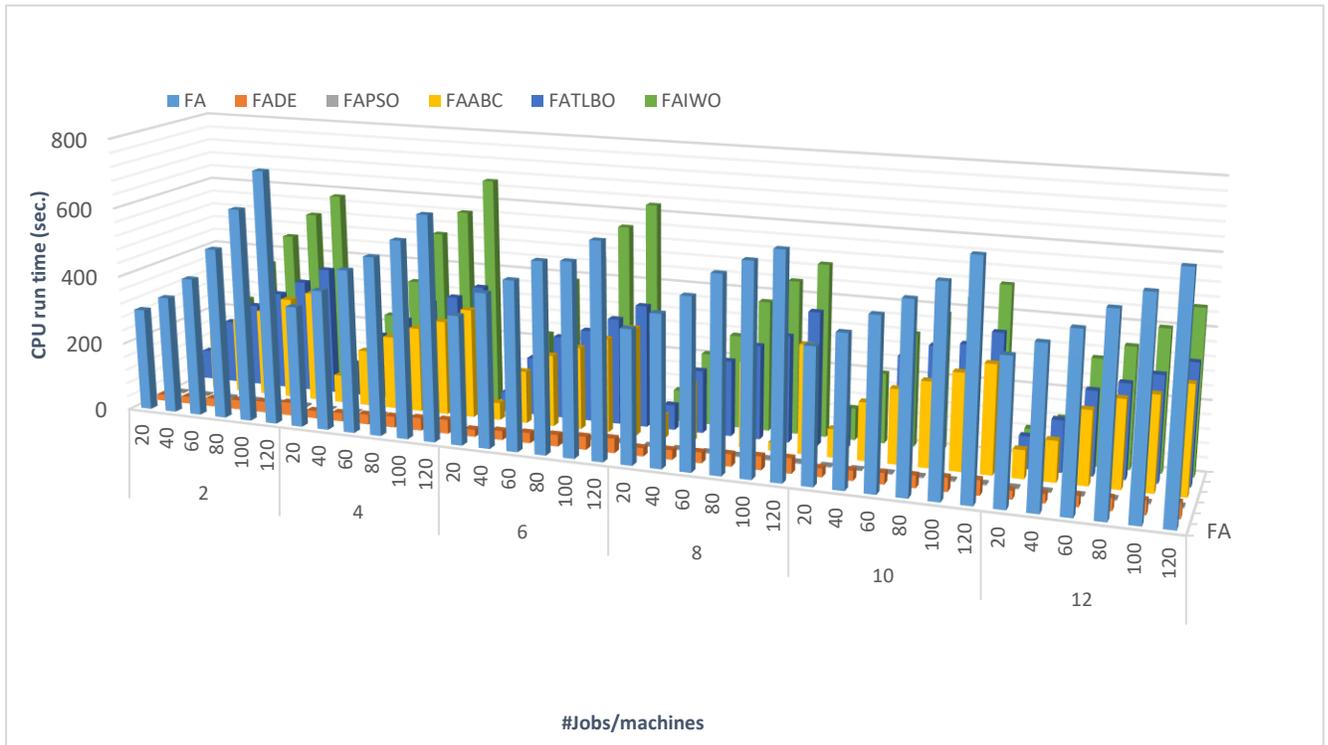

Fig. 13: Average computational time for FA, FAPSO, FADE, FAABC, FATLBO, and FAIWO on all test instances

## 5.4 Contribution of this study to Sustainable Development Goals

The scheduling experiments and results discussed in this paper focus on optimizing the assignment of jobs to machines in industries. The findings highlight the significant importance of this optimization process in reducing the overall makespan of scheduling tasks, which, in turn, contributes to the achievement of Sustainable SDGs. The following are key reasons illustrating the importance of this optimization:

- Enhanced Resource Utilization: Optimizing job assignments ensures efficient utilization of resources within industries. By effectively allocating tasks to machines, idle time is minimized, leading to increased productivity and reduced resource wastage. This optimization aligns with SDG 12 (Responsible Consumption and Production) by promoting sustainable and efficient resource management practices.
- Shortened Production Time: Optimizing the assignment of jobs to machines reduces the overall makespan, resulting in shorter production cycles. This efficiency improvement enhances operational effectiveness, enabling industries to meet customer demands more rapidly and respond to market dynamics swiftly. Consequently, it contributes to SDG 8 (Decent Work and Economic Growth) by fostering productivity and supporting sustainable economic development.
- Improved Customer Satisfaction: By minimizing the makespan through optimized job assignments, industries can deliver products or services to customers more promptly. Reduced lead times enhance customer satisfaction, trust, and loyalty, reinforcing the industry's market position. This alignment with SDG 9 (Industry, Innovation, and Infrastructure) promotes resilient infrastructure and sustainable industrial practices.
- Environmental Footprint Reduction: Efficient job assignment optimization minimizes energy consumption and associated emissions, leading to a reduced environmental impact. Balancing workloads across machines helps avoid energy wastage and prevents overburdening specific resources. Consequently, this optimization supports SDG 13 (Climate Action) by reducing carbon emissions and fostering environmentally responsible operations.
- Resource Conservation and Waste Minimization: Optimal job assignments minimize downtime, machine idling, and unnecessary changeovers, leading to decreased waste generation. By reducing material scrap, energy waste, and water consumption, industries contribute to resource conservation and sustainable waste management practices. This aligns with SDG 12 (Responsible Consumption and Production) by promoting sustainable production methods and the principles of the circular economy.
- Sustainable Production Planning: Optimized job assignments facilitate effective production planning and capacity utilization. Aligning production schedules with demand forecasts helps prevent overproduction and excess inventory accumulation. Consequently, this optimization supports SDG 12 (Responsible Consumption and Production) by promoting sustainable production practices and minimizing waste across the supply chain.

In summary, the optimization of job assignments to machines in industries has a substantial impact on reducing the overall makespan of scheduling tasks. Through enhanced resource utilization, shortened production time, improved customer satisfaction, reduced environmental footprint, resource conservation, and sustainable production planning, this optimization contributes to the achievement of various SDGs. It fosters sustainable development by promoting responsible consumption, economic growth, climate action, and efficient production practices.

## 6. Conclusion and future direction

In conclusion, the application of metaheuristic optimization techniques for unrelated parallel machine scheduling holds great potential in contributing to the achievement of Sustainable Development Goals. This paper has presented a comprehensive analysis and comparison of various metaheuristic algorithms, highlighting their strengths and limitations in addressing the scheduling problem. Through extensive experimentation, it has been observed that hybrid algorithms such as FAIWO, FAABC, and FATLBO exhibit superior performance in terms of convergence, makespan reduction, and search space efficiency. These algorithms have demonstrated their ability to generate near-optimal solutions for large instances of the scheduling problem. Moreover, this research has identified the need for further investigation into enhancing the convergence speed of the FA algorithm. Developing novel techniques specifically tailored to address this challenge would contribute to improving the overall efficiency and effectiveness of FA for unrelated parallel machine scheduling.

Moreover, the study findings highlighted the significant contribution of metaheuristic optimization algorithms in attaining SDGs within the Unrelated Parallel Machines Scheduling Problem context. These algorithms play a vital role by optimizing the assignment of jobs to machines, ultimately reducing the overall makespan of scheduling tasks. This optimization process leads to sustainable and responsible practices in scheduling and resource management. By effectively optimizing the assignments of jobs to machines, these algorithms enhance resource utilization and minimize idle time. This promotes efficient use of resources and reduces waste, aligning with sustainable development goals related to responsible consumption and production. Furthermore, the reduction in the overall makespan achieved through optimization improves operational efficiency and productivity. Shorter makespan means tasks are completed in a shorter time, leading to faster delivery of products or services. This contributes to customer satisfaction and supports sustainable economic growth, in line with sustainable development goals related to decent work and economic growth.

Moreover, by minimizing the makespan, metaheuristic optimization algorithms help reduce energy consumption and associated emissions. This aligns with sustainable development goals related to climate action by promoting energy efficiency and reducing the environmental impact of scheduling operations. Additionally, the optimized scheduling facilitated by these algorithms enhances resource planning and management. By minimizing makespan, organizations can effectively allocate resources, reducing the need for additional machines and optimizing resource usage. This supports sustainable development goals related to responsible resource management and efficient infrastructure.

Future research directions in this domain could explore the integration of machine learning and data-driven approaches into metaheuristic optimization algorithms, enabling them to adapt and learn from problem instances and historical data. Additionally, the incorporation of real-world constraints, such as energy consumption, carbon footprint, and resource utilization, would align the scheduling process with the broader objective of achieving sustainable development. Furthermore, extending the analysis to include dynamic and stochastic scheduling scenarios would provide insights into the performance and adaptability of metaheuristic algorithms in real-time scheduling environments. In conclusion, by advancing the capabilities of metaheuristic optimization for unrelated parallel machine scheduling, we can contribute to the broader agenda of sustainable development while effectively addressing the challenges associated with complex scheduling problems.